\documentclass[runningheads]{llncs}
\usepackage{graphicx}
\usepackage{amsmath,amssymb} 
\usepackage{relsize}
\usepackage{color}
\usepackage{cite}
\usepackage{booktabs}
\usepackage{multirow}
\usepackage{algorithm, algpseudocode}
\usepackage[misc]{ifsym}
\usepackage{autonum}

\newif\ifreview
\reviewfalse

\ifreview
\usepackage[width=122mm,left=12mm,paperwidth=146mm,height=193mm,top=12mm,paperheight=217mm]{geometry}
\usepackage{ruler}
\else
\usepackage[width=122mm,left=16mm,paperwidth=155mm,height=193mm,top=21mm,paperheight=235mm]{geometry}
\fi

\newcommand{\miniparagraph}[1]{\vspace{0.5em}\noindent\textsf{\textbf{#1:~}}}
\newcommand{\secref}[1]{Sec.\,\ref{#1}}
\newcommand{\figref}[1]{Fig.\,\ref{#1}}
\newcommand{\tabref}[1]{Table\,\ref{#1}}

\renewcommand{\eqref}[1]{Eq.\,\ref{#1}}
\newcommand{\mathbar}{\text{-}}

\newcommand{\inclongleftarrow}{\stackrel{+}\longleftarrow}
\newcommand{\shp}{\raisebox{.4ex}{\scriptsize\#}}
\newcommand{\defeq}{\mathrel{\raisebox{0.034em}{$\mathop{:}$}}=}

\algdef{SE}[SUBALG]{Indent}{EndIndent}{}{\algorithmicend\ }%
\algtext*{Indent}
\algtext*{EndIndent}

\makeatletter
\newcommand{\algmargin}{\the\ALG@thistlm}
\makeatother

\algnewcommand{\parState}[1]{\State%
    \parbox[t]{\dimexpr\linewidth-\algmargin}{\strut #1\strut}}

\begin{document}
\pagestyle{headings}
\mainmatter

\title{RSGAN: Face Swapping and Editing using\\ Face and Hair Representation in Latent Spaces} 

\ifreview
\else
\titlerunning{RSGAN: Face Swapping and Editing via\\ Region Separation in Latent Spaces}
\authorrunning{R. Natsume et al.}
\author{Ryota Natsume$^{(\text{\Letter})}$ \and Tatsuya Yatagawa \and Shigeo Morishima}
\institute{Graduate School of Advanced Science and Engineering,\\Waseda University, Tokyo, Japan\\ \email{ryota.natsume.26@gmail.com},~~~ \email{tatsy@acm.org},~~~ \email{shigeo@waseda.jp}}
\fi

\maketitle

\begin{abstract}
    In this paper, we present an integrated system for automatically generating and editing face images through face swapping, attribute-based editing, and random face parts synthesis. The proposed system is based on a deep neural network that variationally learns the face and hair regions with large-scale face image datasets. Different from conventional variational methods, the proposed network represents the latent spaces individually for faces and hairs. We refer to the proposed network as \textit{region-separative generative adversarial network} (RSGAN). The proposed network independently handles face and hair appearances in the latent spaces, and then, face swapping is achieved by replacing the latent-space representations of the faces, and reconstruct the entire face image with them. This approach in the latent space robustly performs face swapping even for images which the previous methods result in failure due to inappropriate fitting or the 3D morphable models. In addition, the proposed system can further edit face-swapped images with the same network by manipulating visual attributes or by composing them with randomly generated face or hair parts.    
\end{abstract}

\section{Introduction}
\label{sec:introduction}

\begin{figure}[tb]
    \centering
    \includegraphics[width=0.85\linewidth]{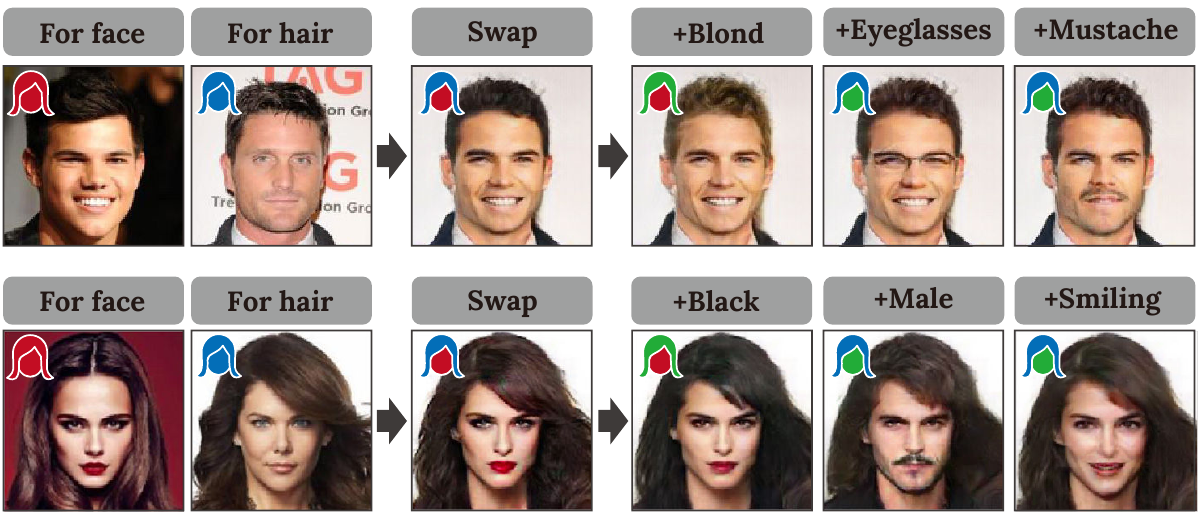}
    \caption{Results of face swapping and additional visual attribute editing with the proposed system. The face regions of images in the first column are embedded to the images in the second column. The face-swapped results are illustrated in the third column, and their appearances are further manipulated by adding visual attributes such as ``blond hair'' and ``eyeglasses''. RSGAN can obtain these results by passing the two input images and visual attributes through the network only once.}
    \label{fig:teaser}
\end{figure}

Human face is an important symbol to recognize individuals from ancient time to the present. Drawings of human faces have been used traditionally to record the human identities of people in authorities. Nowadays, many people enjoy sharing their daily photographs, which usually includes human faces, in social networking websites. In these situations, there has been a potential demand for making the drawings or photographs to be more attractive. As a result of this demand, a large number of studies for face image analysis~\cite{blanz02,cao14,liu15,zhang16_mtcnn} and manipulation~\cite{blanz04,bitouk08,yang11,chai12,kemelmacher16,shu17_tog,fiser17} have been introduced in the research communities of computer graphics and vision.

Face swapping is one of the most important techniques of face image editing that has a wide range of practical applications such as photomontage~\cite{blanz04}, virtual hairstyle fitting~\cite{kemelmacher16}, privacy protection~\cite{bitouk08,mosaddegh14,korshunova16}, and data augmentation for machine learning~\cite{hassner13,mclaughlin15,masi16}. The traditional face swapping methods firstly detect face regions in source and target images. The face region of the source image is embedded into the target image by digital image stitching. To clarify the motivation of our study, we briefly review the previous face-swapping methods in the following paragraphs.

One of the most popular approaches of face swapping is to use the 3D morphable models (3DMM)~\cite{blanz04,nirkin17}. In this class of methods, the face geometries and their corresponding texture maps are first obtained by fitting 3DMM~\cite{blanz02,cao14}. The texture maps of the source and target images are then swapped with the estimated UV coordinates. Finally, the replaced face textures are re-rendered using the estimated lighting condition estimated with the target image. These approaches with the 3DMM can replace the faces even for those with different orientations or in the different lighting conditions. However, these methods are prone to fail the estimations of face geometries or lighting conditions in practice. The incorrect estimations are usually problematic because people can sensitively notice even slight mismatches of these geometries and lighting conditions.

In specific applications of face swapping, such as privacy protection and virtual hairstyle fitting, either of the source image or target image can be selected arbitrarily. For instance, the privacy of the target image can be protected even though the new face region is extracted from a random image. This has suggested an idea of selecting one of the source and target image from large-scale image databases~\cite{bitouk08,kemelmacher16}. The approaches in this class can choose one of two input images such that a selected image is similar to its counterpart. These approaches can consequently avoid replacing faces in difficult situations with different face orientations or different lighting conditions. However, these methods cannot be used for more general purposes of face swapping between arbitrary input face images.

A vast body of recent deep learning research has facilitated face swapping with large-scale image databases. Bao et al. have introduced a face swapping demo in their paper of conditional image generation with their proposed neural network named CVAE-GAN~\cite{bao17}.  Their method uses hundreds of images for each person in the training dataset, and the face identities are learned as the image conditions. A similar technique is used in a desktop software tool ``FakeApp''~\cite{fakeapp} that has recently attracted much attention due to its easy-to-use pipeline for face swapping with deep neural networks (DNN). This tool requires hundreds of images of the two target people for swapping the faces. However, preparing such a large number of portrait images for non-celebrities is rather inadvisable. In contrast to these techniques, Korshunova et al.~\cite{korshunova16} have applied the neural style transfer~\cite{gatys16} to face swapping by fine-tuning the pre-trained network with several tens of images of a single person in a source image. Unfortunately, it is still impractical for most of people to collect many images and to fine-tune the network for generating a single face-swapped image.

In this paper, we address the above problems using a generative neural network that we refer to as ``region-separative generative adversarial network (RSGAN).'' While a rich body of studies for such deep generative models has already been introduced, applying it to face swapping is still challenging. In ordinary generative models, the images or data which a network synthesizes are obtained as training data. However, it is difficult or even impossible to prepare a dataset which includes face images both before and after face swapping because the faces of real people can hardly be swapped without special surgical operations. We tackle this problem by designing the network to variationally learn different latent spaces for each of face and hair regions. A generator network used in the proposed method is trained to synthesize a natural face image from two random vectors that correspond to latent-space representations of face and hair regions. In consequence, the generator network can synthesize a face-swapped image from two latent-space representations calculated from real image samples. The architecture of RSGAN is illustrated in \figref{fig:network}. This architecture consists of two variational autoencoders (VAE) and one generative adversarial network (GAN). The two VAE parts encode face and hair appearances into latent-space representations, and the GAN part generates a natural face image from the latent-space representations of faces and hairs. The detailed description of the network and its training method are introduced in \secref{sec:rsgan}. In addition to face swapping, this variational learning enables other editing applications, such as visual attribute editing and random face parts synthesis. To evaluate face swapping results of the proposed method, we leveraged two metrics of identity preservation and swap consistency. The identity preservation is evaluated using OpenFace~\cite{amos16_openface}, which is an open-source face feature extractor. The consistency of face swapping is evaluated by measuring the absolute difference and multi-scale structural similarity (MS-SSIM)~\cite{wang_msssim} between a input image and a resulting image obtained by swapping faces twice between two input images. The results of applications of RSGAN and their evaluations are shown in \secref{sec:results}.

\miniparagraph{Contributions} As a face swapping and editing system, the proposed method has following advantages over previous methods:
\begin{enumerate}
    \item it provides an integrated system for face swapping and additional face appearance editing;
    \item its applications are achieved by training a single DNN, and it does not require any additional runtime computation such as fine-tuning;
    \item it robustly performs high-quality face swapping even for faces with different face orientations or in different lighting conditions;
\end{enumerate}

\section{Related Work}
\label{sec:related-work}

\subsection{Face swapping}
\label{ssec:face-swapping}

Face swapping has been studied in a number of research for different purposes, such as photomontage~\cite{blanz04}, virtual hairstyle fitting~\cite{kemelmacher16}, privacy protection~\cite{bitouk08,mosaddegh14,korshunova16} and data augmentation for large-scale machine learning~\cite{masi16}. Several studies~\cite{yang11,mosaddegh14} have replaced only parts of the face, such as eyes, nose, and mouth between images rather than swapping the whole face between images. As described in the previous section, one of the traditional approaches for face swapping is based on 3DMM~\cite{blanz04,nirkin17}. Fitting 3DMM to a target face obtains face geometry, texture map and lighting condition approximately~\cite{blanz02,cao14}. Using the 3DMM, face swapping is achieved by replacing texture maps and re-rendering the face appearance using the estimated lighting condition. The main drawback of these 3DMM-based methods is that they require manual alignment of the 3DMM to obtain accurate fitting. To alleviate this problem, Bitouk et al.~\cite{bitouk08} proposed automatic face swapping with a large-scale face image database. Their method first searches a face image with a similar layout to the input image, and then replace the face regions with boundary-aware image composition. A more sophisticated approach was recently proposed by Kemelmacher-Shlizerman~\cite{kemelmacher16}. She carefully designed a handmade feature vector to face image appearances and achieved high-quality face swapping. However, these methods by searching similar images cannot freely select the input images, and are not applicable to arbitrary face image pairs. Recently, Bao et al. have introduced a face swapping demo in their paper of CVAE-GAN~\cite{bao17}, which is a DNN for conditional image generation. In their method, the CVAE-GAN is trained to generate face images of specific people in a training dataset by handling face identities as conditions for generated images. The CVAE-GAN achieves face swapping by changing the conditions of face identities of target images. Korshunova et al. applied neural style transfer, which is another technique of deep learning, to face swapping~\cite{korshunova16}. Their approach is similar to the original neural style transfer~\cite{gatys16} in the sense that a face identity is handled similarly to an artistic style. The face identity of a target face is substituted by that of a source face. The common drawback of these DNN-based models is that users must collect at least dozens of images to obtain a face-swapped image. While collecting such a number of images is possible, it is nevertheless impractical for most of people to collect the images just for their personal photo editing.

\subsection{Face image editing}
\label{ssec:face-editing}

To enhance the visual attractiveness of face images, various techniques, such as facial expression transfer~\cite{liu01,yang11}, attractiveness enhancement~\cite{leyvand08}, face image relighting~\cite{chai15,shu17}, have been proposed in the last decades. In traditional face image editing, underlying 3D face geometries and face parts arrangements are estimated using face analysis tools, such as active appearance models~\cite{cootes01} and 3D morphable models~\cite{blanz02,cao14}. These underlying information are manipulated in editing algorithms to improve attractiveness of output images. On the other hand, recent approaches based on DNNs do not explicitly analyze such information. Typically, an input image and a user's edit intention are fed to an end-to-end DNN, and then, the edit result is directly output from the network. For example, several DNN models~\cite{kingma14,odena16,bao17,choi17} based on autoencoders are used to manipulate visual attributes of faces, in which visual attributes, such as facial expressions and hair colors, are modified to change face image appearances. In contrast,  Brock et al.~\cite{brock16} proposed an image editing system with the paint-based interface in which a DNN synthesizes a natural image output following the input image and paint strokes specified by the users. Several studies for DNN-based image completion~\cite{iizuka17,chen18} have presented demos of manipulating face appearances by filling the parts of an input image with the DNN. However, estimating results of these approaches is rather difficult because they only fill the regions painted by the users such that completed results plausibly exists in training data.

\section{Region-Separative GAN}
\label{sec:rsgan}

\begin{figure*}[tb]
    \centering
    \includegraphics[width=\linewidth]{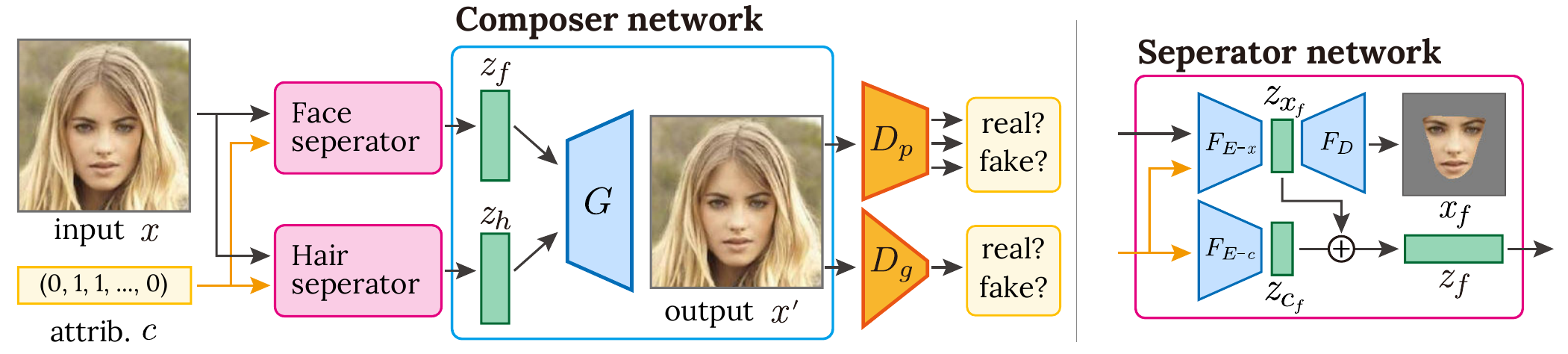}
    \caption{The network architecture of the proposed RSGAN that comprises three partial networks, i.e., two separator networks and a composer network. The separator networks extract latent-space representations $z_f$ and $z_h$ respectively for face and hair regions of an input image $x$. The composer network reconstructs the input face image from the two latent-space representations. The reconstructed image $x'$ and input image $x$ are evaluated by two discriminator networks. The global discriminator $D_g$ distinguishes whether the images are real or fake, and the patch discriminator $D_p$ distinguishes whether local patches of the images are real or fake.}
    \label{fig:network}
\end{figure*}

The main challenge of face swapping with DNNs is the difficulty of preparing face images before and after face swapping because the face of a real person cannot be replaced by that of another person without a special surgical operation. Another possible way to collect such face images is to digitally synthesize them. However, it is an chicken-and-egg problem because the synthesis of such face-swapped images is our primary purpose. To overcome this challenge, we leverage a variational method to represent appearances of faces　and hairs. In face swapping, a face region and a hair region are handled separately in the image space. The face swapping problem is generalized as a problem of composing any pair of face and hair images. The purpose of the proposed RSGAN is to achieve this image composition using latent-space representations of face and hair appearances. In the proposed method, this purpose is achieved by a DNN shown in \figref{fig:network}. As shown in this figure, the architecture of RSGAN consists of two VAEs, which we refer to as \textit{separator network}, and one GAN, which we refer to as \textit{composer network}. In this network, appearances of the face and hair regions are first encoded into different latent-space representations with the separator networks. Then, the composer network generates a face image with the obtained latent-space representations so that the original appearances in the input image is reconstructed. However, training with only latent-space representations from real image samples incurs over-fitting. We found that a RSGAN trained in this way ignores the face representation is the latent space, and synthesizes an image similar to the target image while face swapping. Thus, we also feed random latent-space representations to the composer network such that they are trained to synthesize natural face images rather than over-fitting the training data.

Let $x$ be a training image, and $c$ be its corresponding visual attribute vector. Latent-space representations $z_{x_{f}}$ and $z_{x_{h}}$ of face and hair appearances of $x$ are obtained by a face encoder $F_{E \mathbar x_{f}}$ and a hair encoder $F_{E \mathbar x_{h}}$. Similarly, the visual attribute $c$ is embedded into latent spaces of the attributes. Latent-space representations $z_{c_{f}}$ and $z_{c_{h}}$ of the face and hair attribute vectors are obtained by encoders $F_{E \mathbar c_{f}}$ and $F_{E \mathbar c_{h}}$. As standard VAEs, these latent-space representations are sampled from multivariate normal distributions whose averages and variances are inferred by the encoder networks:
\begin{equation}
z_\ell = \mathcal{N} \left( \mu_\ell, \sigma^2_\ell \right), \quad \left( \mu_\ell, \sigma^2_\ell \right) = F_{E \mathbar \ell}(x, c), \quad \ell \in \{ x_{f}, x_{h}, c_{f}, c_{h} \}, \label{eq:latent-space-sampling}
\end{equation}
where $\mu_\ell$ and $\sigma^2_\ell$ are the average and variance of $z_\ell$ obtained with the encoders. Decoder networks $F_{D \mathbar f}$ and $F_{D \mathbar h}$ for face and hair regions reconstruct the appearances $x_f'$ and $x_h'$ respectively from the corresponding latent-space representation. The composer network $G$ generates the reconstructed appearance $x'$ with the latent-space representations from the encoders. These reconstruction processes are formulated as:
\begin{equation}
x_f' = F_{D \mathbar f}(z_{x_{f}}, z_{c_{f}}), \quad x_h' = F_{D \mathbar h}(z_{x_{h}}, z_{c_{h}}), \quad x' = G(z_{x_{f}}, z_{c_{f}}, z_{x_{h}}, z_{c_{h}}).
\end{equation}
In addition, random variables sampled from a multivariate standard normal distribution $\mathcal{N}(0, 1)$ are used together in the training. Let $\hat{z}_{x_{f}}$, $\hat{z}_{x_{h}}$, $\hat{z}_{c_{f}}$, and $\hat{z}_{c_{h}}$ be the random variables which correspond to $z_{x_{f}}$, $z_{x_{h}}$, $z_{c_{f}}$, and $z_{c_{h}}$, respectively. We also compute a random face image $\hat{x}'$ with these samples:
\begin{equation}
\hat{x}' = G(\hat{z}_{x_{f}}, \hat{z}_{c_{f}}, \hat{z}_{x_{h}}, \hat{z}_{c_{h}}).
\end{equation}
The input image $x$ and two generated images $x'$ and $\hat{x}'$ are evaluated by two discriminator networks $D_g$ and $D_p$. The global discriminator $D_g$ distinguishes whether those images are real or fake as in standard GANs~\cite{goodfellow14}. On the other hand, the patch discriminator $D_p$, which is originally used in an image-to-image network~\cite{isola17}, distinguishes whether local patches from those images are real or fake. In addition, we train a classifier network $C$ to estimate the visual attribute $c^{*}$ from the input image $x$. The classifier network is typically required to edit an image for which visual attributes are not prepared. In addition, the classifier network obtains a visual attribute vector whose entries are in between $0$ and $1$, whereas visual attribute vectors prepared in many public datasets take discrete values of $0$ or $1$. Such intermediate values are advantageous, for example, when we represent dark brown hair with two visual attribute items ``black hair'' and ``brown hair''. Accordingly, we use the estimated attributes $c^{*}$ rather than $c$ even when visual attributes are prepared for $x$.

\subsection{Training}
\label{ssec:training}

In the proposed architecture of RSGAN, three autoencoding processes are performed, each of those reproduces $x_f'$, $x_h'$ and $x'$ from an input image $x$. Following standard VAEs, we define three reconstruction loss functions:
\begin{align}
\mathcal{L}_{rec \mathbar f} &= \mathbb{E}_{x, x_f \sim P_{data}} \big[ \| x_f - x_f' \|_1  \big], \label{eq:face-rec-loss} \\
\mathcal{L}_{rec \mathbar h} &= \mathbb{E}_{x, x_h \sim P_{data}} \big[ \left\| \left( 1 - \beta M_{BG} \right) \odot \left( x_h - x_h' \right) \right\|_1  \big], \label{eq:hair-rec-loss} \\
\mathcal{L}_{rec} &= \mathbb{E}_{x \sim P_{data}} \big[  \left\| \left(1 - \beta M_{BG} \right) \odot (x - x') \right\|_1  \big], \label{eq:rec-loss}
\end{align}
where $M_{BG}$ is a background mask which take 0 for foreground pixels and 1 for background pixels, and an operator $\odot$ denotes per-pixel multiplication. The background mask $M_{BG}$ is used to train the network to synthesize more detailed appearances in foreground regions. In our implementation, we used a parameter $\beta = 0.5$ to halve the least square errors in the background. The Kullback Leibler loss function is also defined as standard VAEs for each of the four encoders:
\begin{equation}
\mathcal{L}_{KL \mathbar \ell} = \frac{1}{2} \left(  \mu_\ell^T \mu_\ell + \sum \left( \sigma_\ell - \log (\sigma_\ell) - 1 \right) \right),　\qquad \ell \in \{ x_{f}, x_{h}, c_{f}, c_{h}  \}. \label{eq:kl-losses}
\end{equation}
The set of separator and composer networks, and the two discriminator networks are trained adversarially as standard GANs. Adversarial losses are defined as:
\begin{align}
\mathcal{L}_{adv \mathbar \ell} =& - \mathbb{E}_{x \sim P_{data}} \left[ \log D_\ell (x) \right] \nonumber \\
& - \mathbb{E}_{z \sim P_z} \left[ \log (1\! - \! D_\ell (x')) \right] \nonumber \\
& - \mathbb{E}_{z \sim P_z} \left[ \log (1\! - \! D_\ell (\hat{x}')) \right], \quad \ell \in \{ g, p \}. \label{eq:adv-loss}
\end{align}
In addition, the classifier network $C$ is trained to estimate correct visual attributes $c^{*}$ which is close to $c$. We defined a cross entropy loss function $L_{BCE}$, and used it to define a loss function of classifier network $\mathcal{L}_C$:
\begin{equation}
\mathcal{L}_{C} = L_{BCE}(c, c^{*}) \defeq - \sum_{i} \left( c_i \log c_i^{*} + (1 - c_i) \log (1 - c_i^{*}) \right),
\label{eq:classifier-loss}
\end{equation}
where $c_i$ denotes the $i$-th entry of the visual attribute vector $c$. To preserve the visual attributes in generated images $x'$ and $\hat{x}'$, we add the following loss functions to train the composer network:
\begin{equation}
\mathcal{L}_{GC} = L_{BCE}(c, c') + L_{BCE}(c, \hat{c}'),
\label{eq:gen-classifier-loss}
\end{equation}
where $c'$ and $\hat{c}'$ are estimated visual attributes of $x'$ and $\hat{x}'$, respectively.

The total loss function for training RSGAN is defined by a weighted sum of the above loss functions:
\begin{align}
\mathcal{L} &= \lambda_{rec} (\mathcal{L}_{rec \mathbar f} + \mathcal{L}_{rec \mathbar h} + \mathcal{L}_{rec}) \nonumber \\
& + \lambda_{KL} (\mathcal{L}_{KL \mathbar x_{f}} + \mathcal{L}_{KL \mathbar x_{h}} + \mathcal{L}_{KL \mathbar c_{f}} + \mathcal{L}_{KL \mathbar c_{h}}) \nonumber \\
& + \lambda_{adv \mathbar g} \mathcal{L}_{adv \mathbar g} + \lambda_{adv \mathbar p} \mathcal{L}_{adv \mathbar p} \\
& + \lambda_C \mathcal{L}_C + \lambda_{GC} \mathcal{L}_{GC}
\end{align}
We empirically determined the weighting factors as $\lambda_{rec} = 4000$, $\lambda_{KL} = 1$, $\lambda_{adv \mathbar g} = 20$, $\lambda_{adv \mathbar p} = 30, \lambda_{C} = 1$, and $\lambda_{GC} = 50$. In our experiment, the loss functions were minimized using ADAM optimizer~\cite{kingma14_adam} with an initial learning rate of 0.0002, $\beta_1 = 0.5$, and $\beta_2 = 0.999$. The size of a mini-batch was 50. The detailed training algorithm is provided in the supplementary materials.

\begin{figure}[tb]
    \centering
    \includegraphics[width=0.8\linewidth]{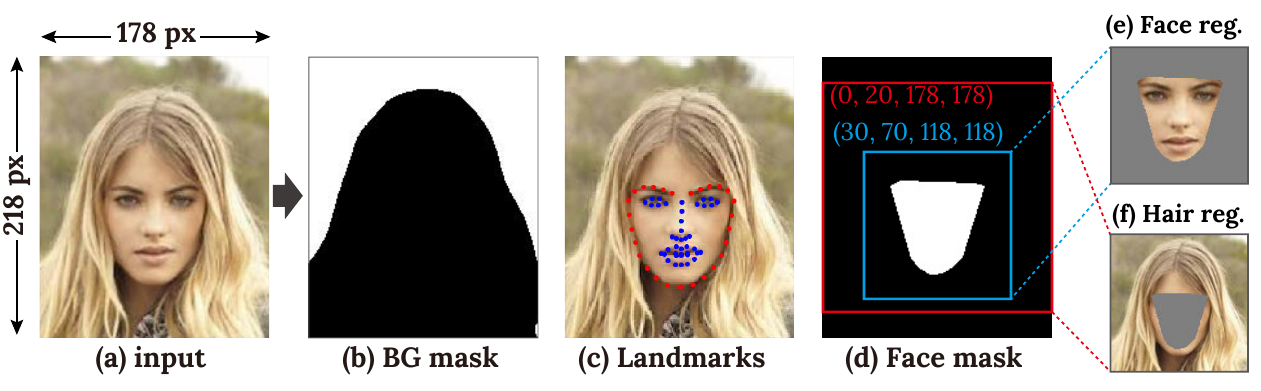}
    \caption{The process of generating face and hair region images from portraits in CelebA~\cite{liu15}. The background mask in (b) is computed with PSPNet~\cite{zhao16_pspnet}, which is a state-of-the-art DNN-based semantic segmentation. Clipping rectangles in (d) for face and hair regions are computed using blue facial landmarks in (c). To improve the reconstruction quality of face identities, face regions are magnified with larger scale than hair regions.}
    \label{fig:dataset}
\end{figure}

\subsection{Dataset}
\label{ssec:dataset}

The training of RSGAN requires to sample face image $x$, face region image $x_f$, hair region image $x_h$, and background mask $M_{BG}$ from real samples. For this purpose, we computationally generate face and hair region images with a large-scale face image dataset, i.e., CelebA~\cite{liu15}. Figure\,\ref{fig:dataset} illustrates the process of dataset generation. The size of original images in CelebA is $178 \times 218$ (\figref{fig:dataset}(a)). We first estimate the foreground mask using PSPNet~\cite{zhao16_pspnet}, which is a state-of-the-art semantic segmentation method, with the ``person'' label. The background mask is obtained by inverting masked and non-masked pixels in the foreground mask (\figref{fig:dataset}(b)). Second, we extract 68 facial landmarks (\figref{fig:dataset}(c)) with a common machine learning library, i.e., Dlib~\cite{king09}. The face region is defined with the 41 landmarks that correspond to eyes, nose, and mouth, which are indicated with blue circles in \figref{fig:dataset}(c). We calculate a convex hull of these landmarks and stretch the hull by 1.3 times and 1.4 times along horizontal and vertical directions, respectively. The resulting hull is used as a face mask as depicted in \figref{fig:dataset}(d). The face and hair regions are extracted with the mask and crop these regions to be square (\figref{fig:dataset}(e) and (f)). The face region has its top-left corner at $(30, 70)$ and its size is $118 \times 118$. The hair region has its top-left corner at $(0, 20)$ and its size is $178 \times 178$.  Finally, we resize these cropped images to the same size. In our experiment, we resize them to $128 \times 128$. Since the face region is more important to identify a person in the image, we used a higher resolution for the face region. While processing images in the dataset, we could properly extract the facial landmarks for 195,361 images out of 202,599 images included in CelebA. Among these 195,361 images, we used 180,000 images for training and the other 15,361 images for testing.

\subsection{Face swapping with RSGAN}
\label{ssec:how-to-face-swap}

A face-swapped image is computed from two images $x_1$ and $x_2$ with RSGAN. For each of these images, visual attributes $c_1^{*}$ and $c_2^{*}$ are first estimated by the classifier. Then, latent-space representations of these variables $z_{1, x_{f}}$, $z_{1, c_{f}}$, $z_{2, x_{h}}$, and $z_{2, c_{h}}$ are computed by the encoders. Finally, the face-swapped image is generated by the composer network as $x' = G(z_{1, x_{f}}, z_{1, c_{f}}, z_{2, x_{h}}, z_{2, c_{h}})$. This operation of just feeding two input images to RSGAN usually performs face swapping appropriately. However, hair and background regions in an input image are sometimes not recovered properly with RSGAN. To alleviate this problem, we optionally perform gradient-domain image stitching for a face-swapped image. In this operation, the face region of a face-swapped image is extracted with a face mask, which is obtained in the same manner as in the dataset generation. Then, the face region of the face-swapped image is composed of the target image by a gradient-domain image composition~\cite{levin04}. In order to distinguish these two approaches, we denote them as ``RSGAN'' and ``RSGAN-GD'', respectively. Unless otherwise specified, the results shown in this paper are computed only using RSGAN without the gradient-domain stitching.

\section{Results and Discussion}
\label{sec:results}

\begin{figure*}[tb]
    \centering
    \includegraphics[width=0.9\linewidth]{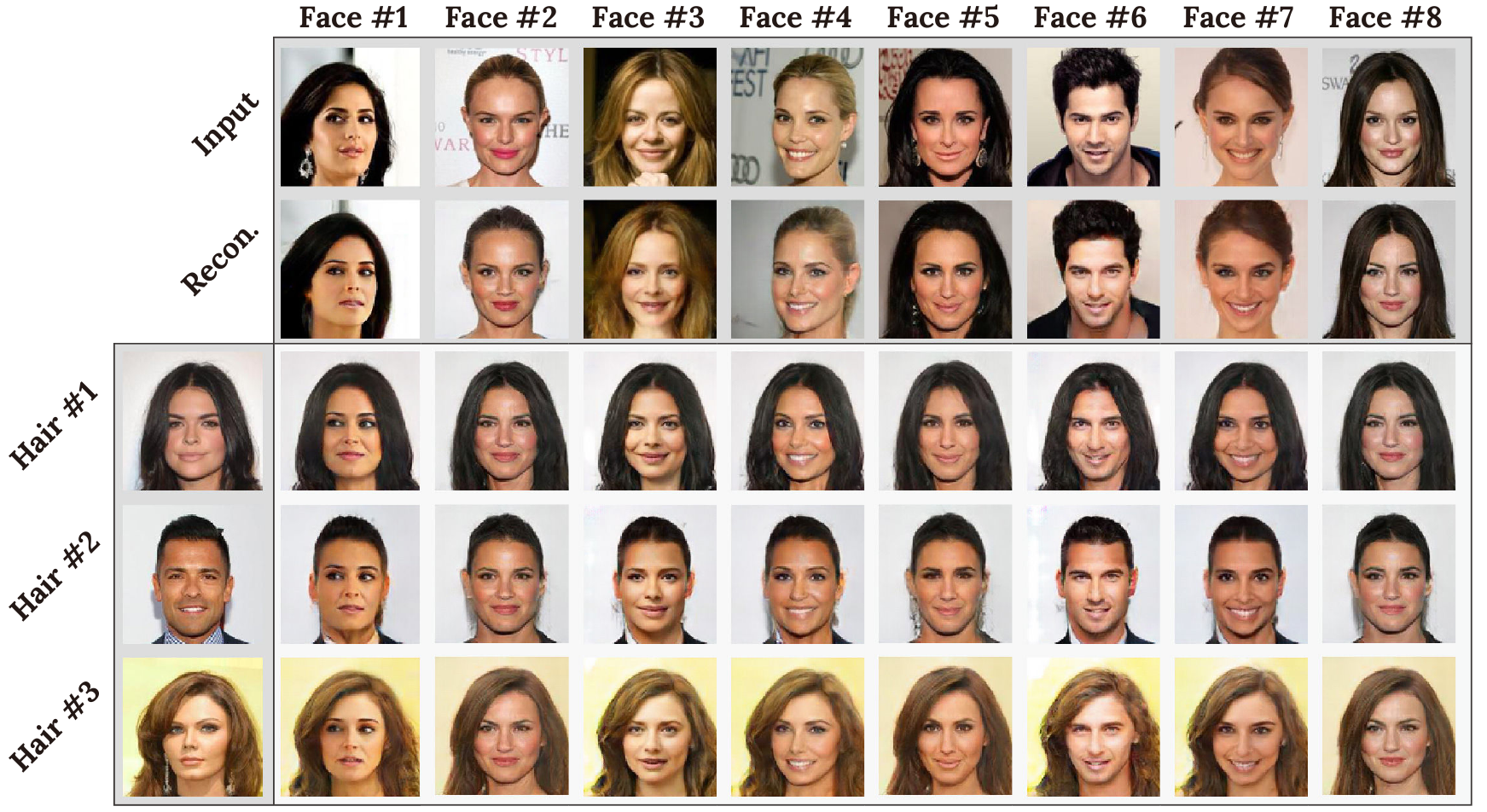}
    \caption{Face swapping results for different face and hair appearances. Two top rows in this figure represent the original inputs and their reconstructed appearances obtained by RSGAN. In these results, face regions of the images in each row is replaced by a face from the image in each column.}
    \label{fig:results-swapping}
\end{figure*}

\begin{figure}[tb]
    \centering
    \includegraphics[width=0.55\linewidth]{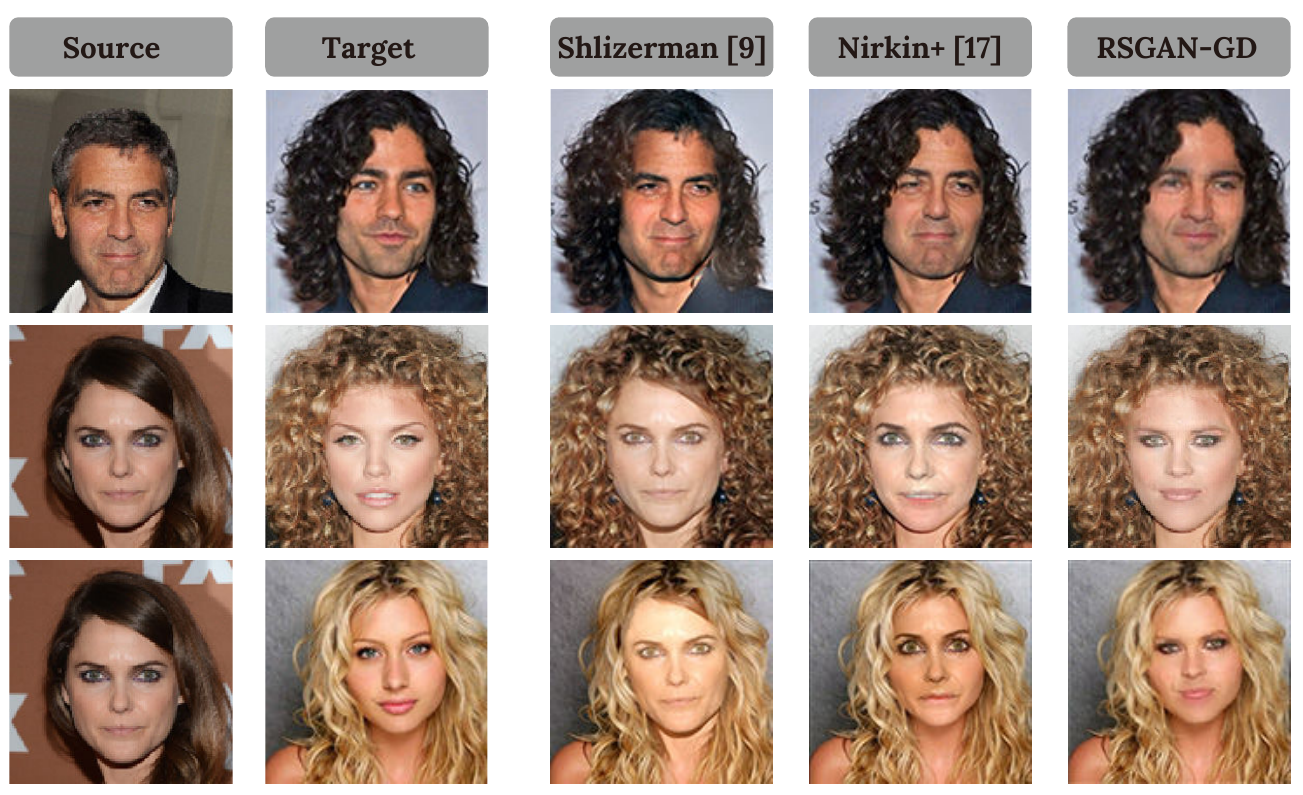}
    \caption{Comparisons to the state-of-the-art face swapping methods~\cite{kemelmacher16,nirkin17}.}    
    \label{fig:comparison}
\end{figure}

This section introduces the results of applications using the pre-trained RSGAN. For these results, we implemented a program using TensorFlow~\cite{abadi2016_tensorflow} in Python, and executed it on a computer with an Intel Xeon 3.6 GHz E5-1650 v4 CPU, NVIDIA GTX TITAN~X GPU, and 64 GB RAM.  We used 180,000 training images, and trained the proposed RSGAN network over 120,000 global steps. The training spent about 50 hours using a single GPU. All the results in this paper are generated using test images which are not included in the training images.

\miniparagraph{Face swapping}
Face-swapping results of the proposed system are illustrated in \figref{fig:results-swapping}. In this figure, the first row illustrates source images, the second row illustrates reproduced appearances for the source images, and the bottom three rows illustrate face-swapped results for different target images in the leftmost column. In each result, we observe that face identities, expressions, shapes of facial parts, and shading are naturally presented in face-swapped results. Among these input image pairs, there are a large difference in the facial expression between Face \shp 7 and Hair \shp 1, a difference in the face orientation between Face \shp 4 and Hair \shp 2, and a difference in the lighting condition in Face \shp 3 and Hair  \shp 1. Even for such input pairs, the proposed method achieves natural face swapping. In addition, we compared our face-swapping results with the state-of-the-art methods~\cite{kemelmacher16,nirkin17} in \figref{fig:comparison}. The results are compared for the input images used in~\cite{kemelmacher16} that are searched from a large-scale database such that their layouts are similar to the source images. While these input images are more favorable for~\cite{kemelmacher16}, the results of our RSGAN-GD are compatible to those of~\cite{kemelmacher16}. Compared to the other state-of-the-art method~\cite{nirkin17}, the sizes of facial parts in our results look more natural in the sense that the proportions of the facial parts to the entire face sizes are more similar to those in the source images. As reported in the paper~\cite{nirkin17}, these performance losses are due to their sensitiveness to the quality of landmark detection and 3DMM fitting, even though their proposed semantic segmentation is powerful.

\begin{figure}[tb]
    \vspace*{2em}
    \centering
    \includegraphics[width=0.95\linewidth]{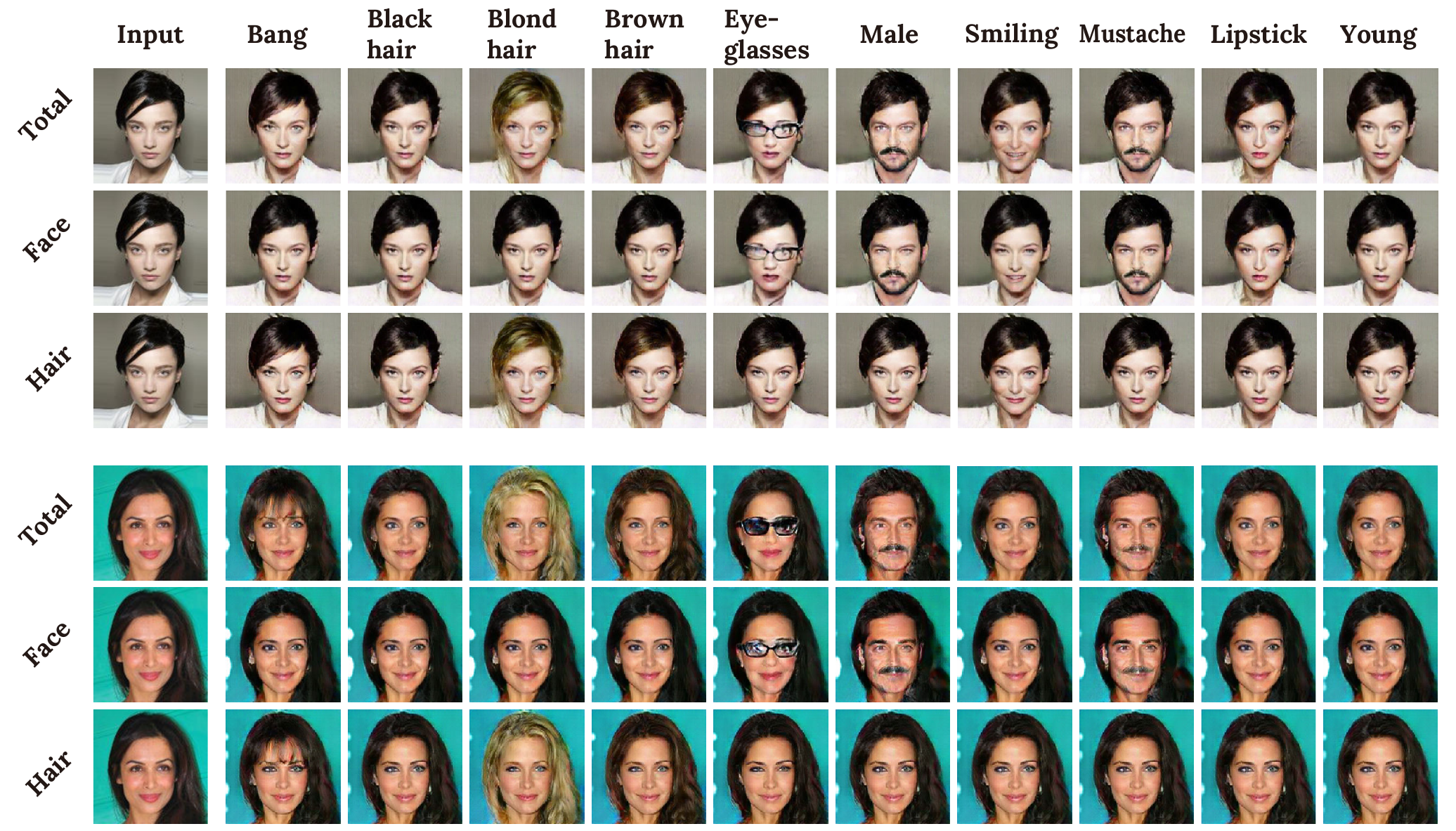}
    \caption{Results of visual attribute editing using RSGAN. In this figure, the results in the rows marked with ``Total'' are obtained adding a new visual attribute both for face and hair regions. The visual attributes used for these results are indicated in the top. On the other hand, the results in the rows marked with ``Face'' are obtained by adding a new visual attribute only for the face region, and the original visual attributes are used for the hair region. The results in the rows marked with ``Hair'' are generated in the same way.}
    \label{fig:results-attrs}
\end{figure}

\miniparagraph{Visual attribute editing}
To perform face swapping together with visual attribute vectors, the proposed RSGAN embeds visual attribute vectors into the latent spaces of face and hair visual attributes. As a result, the proposed editing system enables to manipulate the attributes in only either of face or hair region. The results of visual attribute editing are illustrated in \figref{fig:results-attrs}. This figure includes two image groups, each of which has three rows. In the first row, visual attributes indicated on the top is added to both face and hair regions. In the second and third rows, the visual attributes are added to either of the face or hair region. As shown in this figure, adding visual attributes for only one of face and hair region do not affect the other region. For example, the hair colors have not been changed when the attribute ``Blond hair'' is added to the face regions. In addition, the attributes such as ``Male'' and ``Aged'', which can affect both regions, change the appearance of only one region when they are added to either of two regions. For example, the attribute ``Male'' is added to the face regions, only face appearances become masculine while hair appearances are not changed. In addition, the visual attribute editing can be applied to the face-swapped images with RSGAN. The results for this application is shown in \figref{fig:teaser}. Note that RSGAN can achieve both face swapping and visual attribute editing by feeding two input images and modified visual attribute vectors to the network at the same time.

\begin{figure}[tb]
    \centering
    \includegraphics[width=0.8\linewidth]{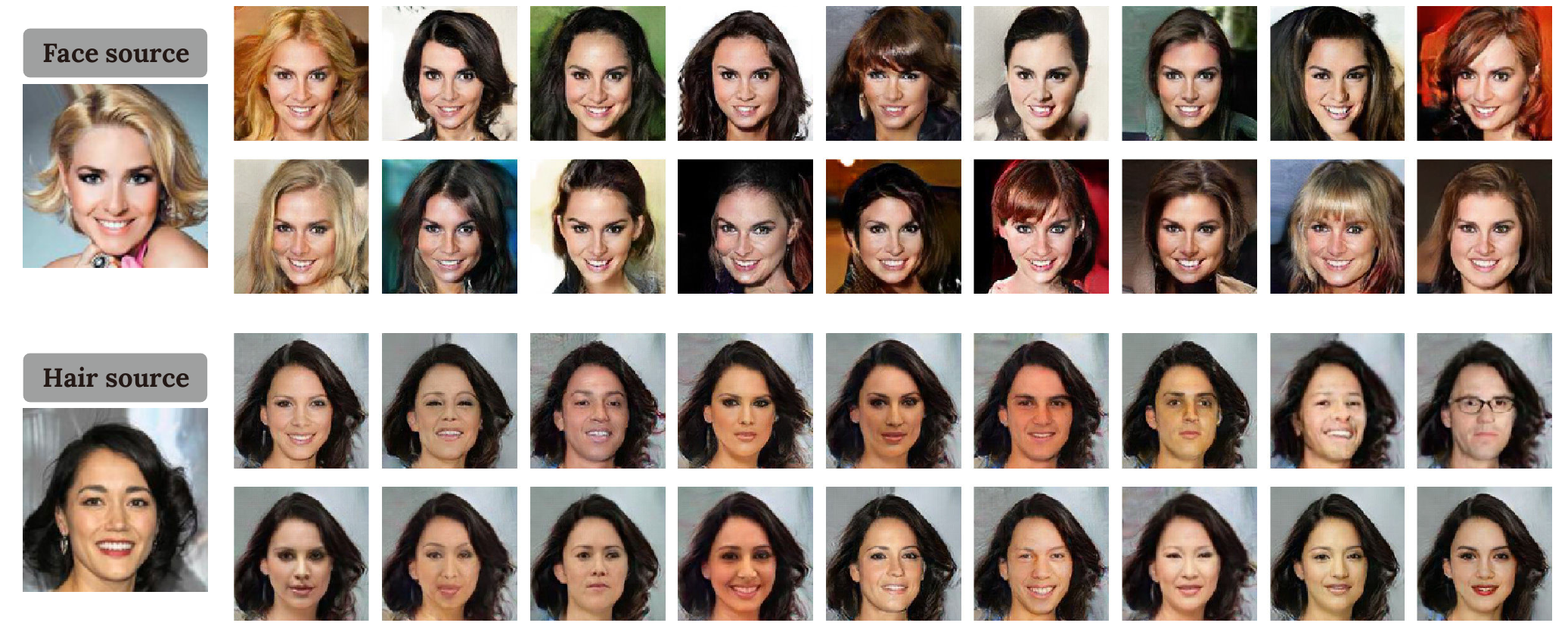}
    \caption{Results of random face and hair parts generation and composition. We can sample independent latent-space representations for face and hair appearances, and combine them with the proposed RSGAN. In the top group, random hair appearances are combined with the face region of an input image in the left. In the bottom group, random face appearances are combined with the hair region of an input image.}
    \label{fig:results-sampling}
\end{figure}

\miniparagraph{Random face parts synthesis}
With the proposed RSGAN, we can generate a new face image which has an appearance of face or hair in a real image sample, and an appearance of the counterpart region defined by a random latent-space. Such random image synthesis is used in privacy protection by changing face regions randomly, and in data augmentation for face recognition by changing hair regions randomly. The results for the random image synthesis are shown in \figref{fig:results-sampling}. This figure consists of two group of images in the top and bottom. In each group, an input image is shown in the left. Its face or hair region is combined with random hair or face region in the right images. The top group illustrates the images with random hairs, and the bottom group illustrates those with random faces. Even though the random face and hair regions cover a significant range of appearances, the appearances of face and hair in the real inputs are preserved appropriately in the results. 

\begin{figure}[tb]
    \centering
    \includegraphics[width=0.8\linewidth]{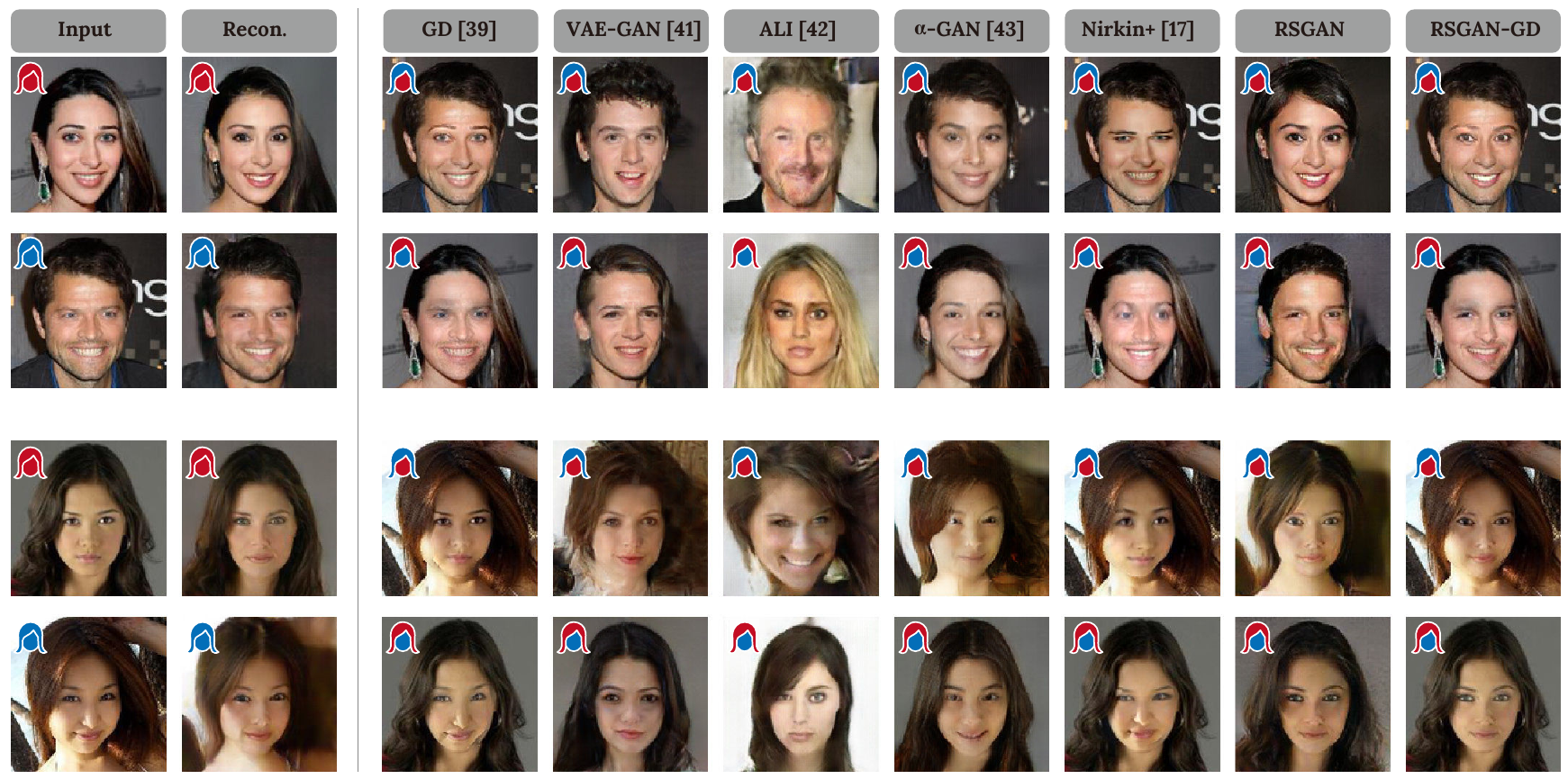}
    \caption{Face swapping results of the proposed RSGAN and the other methods compared in \tabref{tab:experiments}. In two image groups in the top and bottom, face regions of the two input images in the leftmost column are replaced.}
    \label{fig:swap-other-gans}
\end{figure}

\begin{table}[tb]
    \scriptsize
    \centering
    \caption{Performance evaluation in identity preservation and swap consistency.}
    \label{tab:experiments}
    \begin{tabular*}{\linewidth}{l@{\extracolsep{\fill}}llllll}
        \toprule
        & & OpenFace & \multicolumn{2}{c}{Abs. Errors} & \multicolumn{2}{c}{MS-SSIM} \\
        \cmidrule{3-3} \cmidrule{4-5} \cmidrule{6-7}
        & & Swap & Recon. & Swap $\times 2$  & Recon. & Swap $\times 2$ \\
        \midrule
        \multirow{2}{*}{VAE-GAN~\cite{larsen_vaegan}} & Avg. & 1.598 & 0.082 & 0.112 & 0.694 & 0.563 \\
                                                                                                       & Std. & 0.528 & 0.018 & 0.024 & 0.089 & 0.099 \\
        \midrule
        \multirow{2}{*}{ALI~\cite{dumoulin16}} & Avg. & 1.687 & 0.230 & 0.270 & 0.338 & 0.254 \\
                                                                                       & Std. & 0.489 & 0.065 & 0.068 & 0.133 & 0.108 \\
        \midrule
        \multirow{2}{*}{$\alpha$-GAN~\cite{rosca17}} & Avg. & 1.321 & \textbf{0.058} & 0.099 & \textbf{0.823} & 0.638 \\
                                                                                                   & Std. & 0.465 & 0.013 & 0.026 & 0.057 & 0.102  \\
        \midrule
        \multirow{2}{*}{Nirkin et al.~\cite{nirkin17}} & Avg. & \textbf{0.829} & --- & \textbf{0.027} & --- & \textbf{0.961} \\
                                                                                                & Std. & 0.395 & --- & 0.010 & --- & 0.022  \\
        \midrule
        \multirow{2}{*}{\textbf{RSGAN}} & Avg. & \textit{1.127} & \textit{0.069} & \textit{0.093} & \textit{0.760} & \textit{0.673} \\
                                                                          & Std. & 0.415 & 0.016 & 0.020 & 0.074 & 0.087  \\
        \bottomrule
    \end{tabular*}
\end{table}

\subsection{Experiments}
\label{ssec:experiments}

We evaluated the face swapping results of the proposed and other previous methods using two metrics, i.e., identity preservation and swap consistency. In this experiment, we compared these two values with previous self-reproducing generative networks VAE-GAN~\cite{larsen_vaegan}, ALI~\cite{dumoulin16}, $\alpha$-GAN~\cite{rosca17} and the state-of-the-art face swapping method by Nirkin et al.~\cite{nirkin17}. With the generative networks, we computed face-swapped results in three steps. First, we compute a face mask in the same manner as in our dataset synthesis. Second, the face region of the source image in the mask is copy-and-pasted to the target image such that the two eye locations are aligned. Finally, the entire image appearance after copy-and-pasting is repaired by feeding it to self-reproducing networks. The examples of the face-swapped images made by these algorithms are illustrated in \figref{fig:swap-other-gans}. We computed the results of different algorithms for 1,000 random image pairs selected from 15,361 test images. The averages and standard deviations of the two metrics are provided in \tabref{tab:experiments}. In this table, the best score in each column is indicated with bold characters, and the second-best score is indicated with italic characters.

The identity preservation in face swapping is evaluated by the squared Euclidean distance between feature vectors of the input and face-swapped images. The feature vectors are computed with OpenFace \cite{amos16_openface}, which is an open-source face feature extractor. The measured distances in the third column indicate that RSGAN outperform the other generative neural networks but it performs worse than Nirkin et al.'s method. However, the method of Nirkin et al. could perform face swapping only 81.7\% of 1,000 test image pairs used in this experiment because it often fails to fit the 3DMM to at least one of the two input images. In contrast, the proposed RSGAN and the other method based on generative neural networks perform face swapping for all the test images. Therefore, we consider face swapping by RSGAN is practically useful even though the identity preservation is slightly worse than the state-of-the-art method of Nirkin et al. 

The swap consistency is evaluated with an absolute difference and MS-SSIM~\cite{wang_msssim} between an input image and a resulting image obtained after swapping the face region twice between two input images. For the previous generative neural networks and RSGAN, we computed these values also for images reconstructed by the networks. As shown in \tabref{tab:experiments}, evaluation results with absolute errors and MS-SSIM indicate that the method of Nirkin et al. outperforms the generative neural networks including RSGAN. We consider this is because Nirkin et al.'s method generates only a face region while face swapping, whereas the generative neural networks synthesize both the face and hair regions. Therefore, the scores of absolute differences and MS-SSIM becomes relatively lower for the method of Nirkin et al. in which the differences in pixel values only occur in the face regions. In addition, Nirkin et al.'s method is rather unstable for using in practice as mentioned in the previous paragraph. Consequently, the proposed method with RSGAN is worth using in practice because it has achieved the best swap consistency compared to the other generative neural networks as can be seen in the ``Swap $\times$2'' columns.

\subsection{Discussion}
\label{ssec:discussion}

\miniparagraph{Variational vs non-variational} 
For the purpose of visual feature extraction, many variants of autoencoders have been used~\cite{kingma14,gregor15_draw,yang17,li18_agegan}. Among these approaches, non-variational approaches are often preferably used when a real image appearance needs to be reproduced in their applications. For example, recent studies~\cite{yang17,li18_agegan}, which have introduced a similar idea to our study, used non-variational approaches for image parts extraction~\cite{yang17} and manipulation of people's ages in portraits~\cite{li18_agegan}. We have also experimented non-variational one of the proposed RSGAN in a prototype implementation, and we found that its self-reproducibility is slightly better than the variational one that is introduced in this paper. However, considering the wide applicability of the variational one of RSGAN such as random face parts sampling, we determined that the variational one is practically more useful than the non-variational one. 

\miniparagraph{Region memorization with RNN}
In image parts synthesis, some of the previous studies have applied the recurrent neural networks to memorize which parts are already synthesized or not~\cite{gregor15_draw,kwak16,yang17}. Following these studies, we have experimentally inserted the long-short term memory (LSTM)~\cite{hochreiter97_lstm} such that the outputs from the two image encoder networks are fed to it. However, in our experiment, we found that this application of the LSTM makes the training difficult and its convergence slower. The visual qualities of the results in face-swapping and the other applications are not significantly better than RSGAN without LSTM. We illustrated the RSGAN architecture with LSTM, and the results of this network in the supplementary materials.

\miniparagraph{Limitation}
The main drawback of the proposed system is its limited image resolution. In our implementation, the image size of in a training dataset is $128 \times 128$. Therefore, the image editing can be performed only in this resolution. To improve the image resolution, we need to train the network with higher-resolution images as in CelebA-HQ~\cite{karras17}. In recent studies~\cite{karras17,chen18}, training with such a high-resolution image dataset is robustly performed by progressively increasing the resolutions of input images. This approach can be straightforwardly applied to the proposed RSGAN as well. Therefore, the limited image resolution of the proposed system will be evidently resolved.

\section{Conclusion}
\label{sec:Conclusion}

This paper proposed an integrated editing system for face images using a novel generative neural network that we refer to as RSGAN. The proposed system achieves high-quality face swapping, which is the main scope of this study, even for faces with different orientations and in different lighting conditions. Since the proposed system can encode the appearances of faces and hairs into underlying latent-space representations, the image appearances can be modified by manipulating the representations in the latent spaces. As a deep learning technique, the success of the RSGAN architecture and our training method implies that deep generative models can obtain even a class of images that are not prepared in a training dataset. We believe that our experimental results provide a key for generating images which are hardly prepared  in a training dataset.

\ifreview
\else
\section*{Acknowledgments}

This study was granted in part by the Strategic Basic Research Program ACCEL of the  Japan Science and Technology Agency (JPMJAC1602). Tatsuya Yatagawa was supported by a Research Fellowship for Young Researchers of Japan's Society for the Promotion of Science (16J02280). Shigeo Morishima was supported by a Grant-in-Aid from Waseda Institute of Advanced Science and Engineering. The authors would also like to acknowledge NVIDIA Corporation for providing their GPUs in the academic GPU Grant Program.

\fi

\bibliographystyle{splncs}
\bibliography{egbib}


\clearpage
\newpage

\appendix
\renewcommand\thefigure{\thesection\arabic{figure}} 
\renewcommand\thealgorithm{\thesection\arabic{algorithm}}

\setcounter{page}{1}

\title{Supplementary Materials: \\RSGAN: Face Swapping and Editing via\\ Region Separation in Latent Spaces} 

\ifreview
\else
\titlerunning{RSGAN: Face Swapping and Editing via\\ Region Separation in Latent Spaces}
\authorrunning{R. Natsume et al.}
\author{Ryota Natsume$^{1(\text{\Letter})}$, Tatsuya Yatagawa$^{2}$, and Shigeo Morishima$^3$}
\institute{Graduate School of Advanced Science and Engineering,\\Waseda University, Tokyo, Japan\\ $^1$\email{ryota.natsume.26@gmail.com}, $^2$\email{tatsy@acm.org}, $^3$\email{shigeo@waseda.jp}}
\fi

\maketitle

\section{Table of contents}

\begin{figure}[p]
    \centering
    \includegraphics[width=\linewidth]{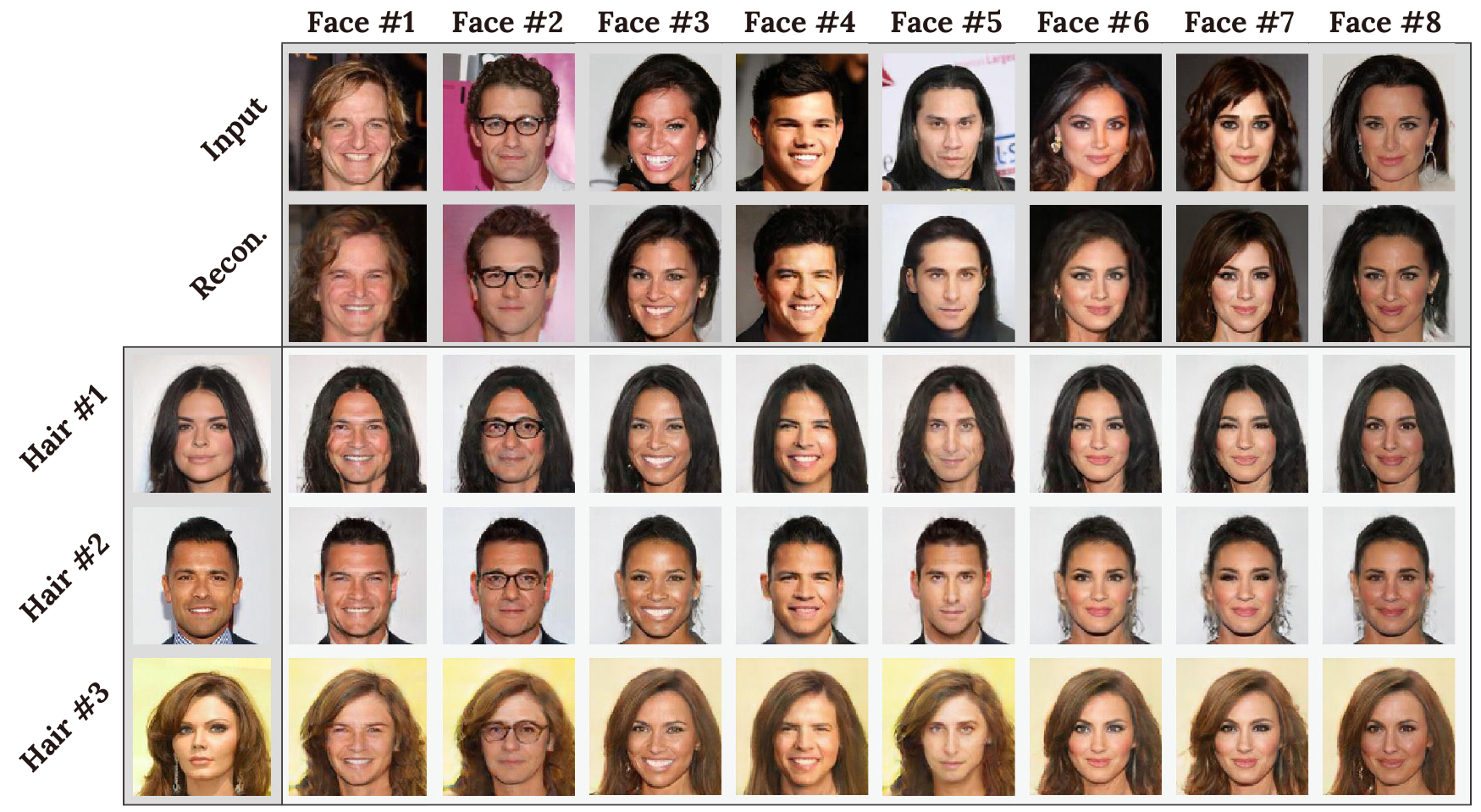} \\
    \vspace{5mm}
    \includegraphics[width=\linewidth]{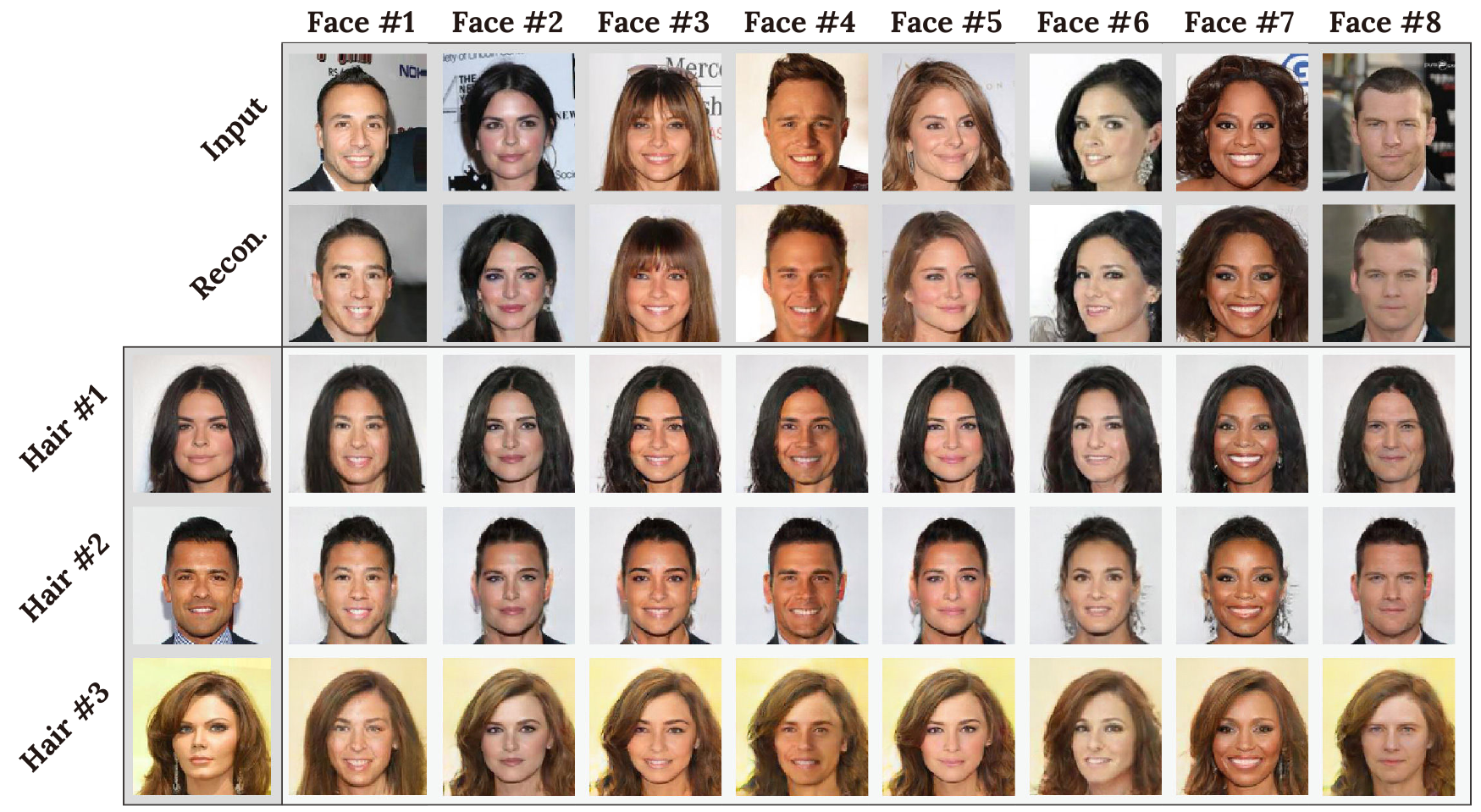}
    \caption{Additional results for face swapping.}
    \label{fig:additional-face-swap}
\end{figure}

\begin{figure}[p]
    \centering
    \includegraphics[width=\linewidth]{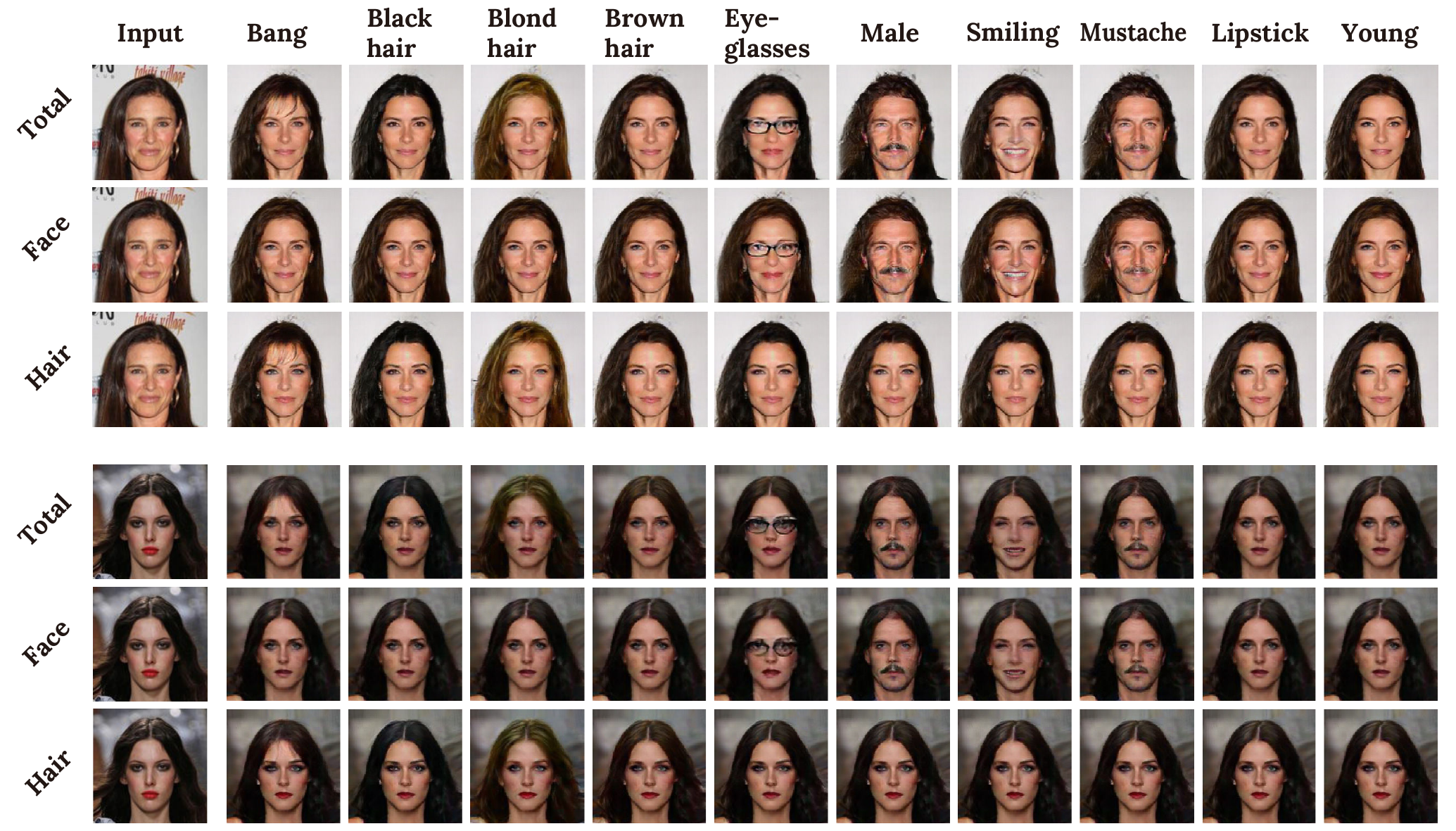} \\
    \vspace{5mm}
    \includegraphics[width=\linewidth]{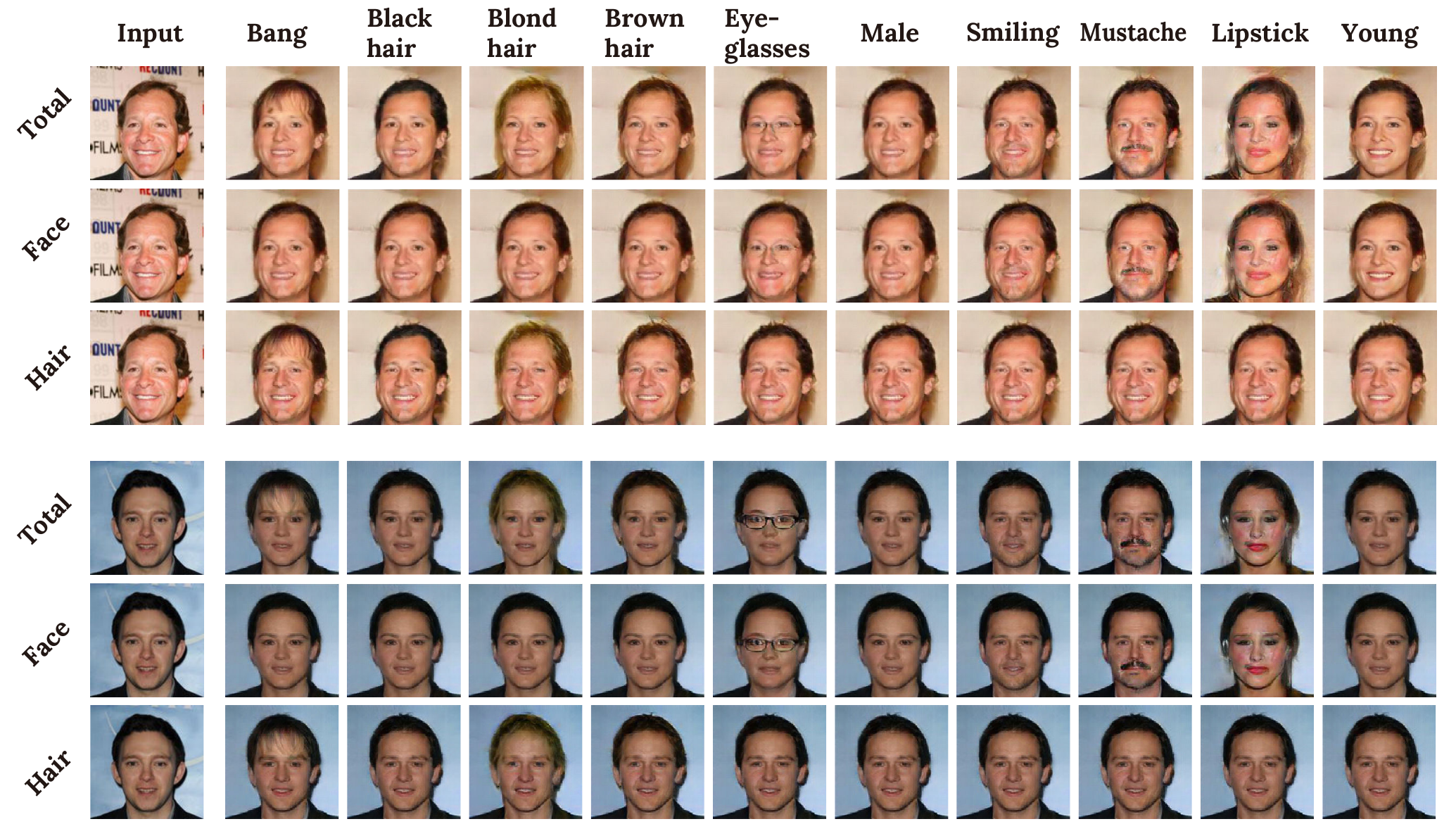}
    \caption{Additional results for visual attribute editing.}
    \label{fig:additional-attr-edit}
\end{figure}

\begin{figure}[p]
    \centering
    \includegraphics[width=\linewidth]{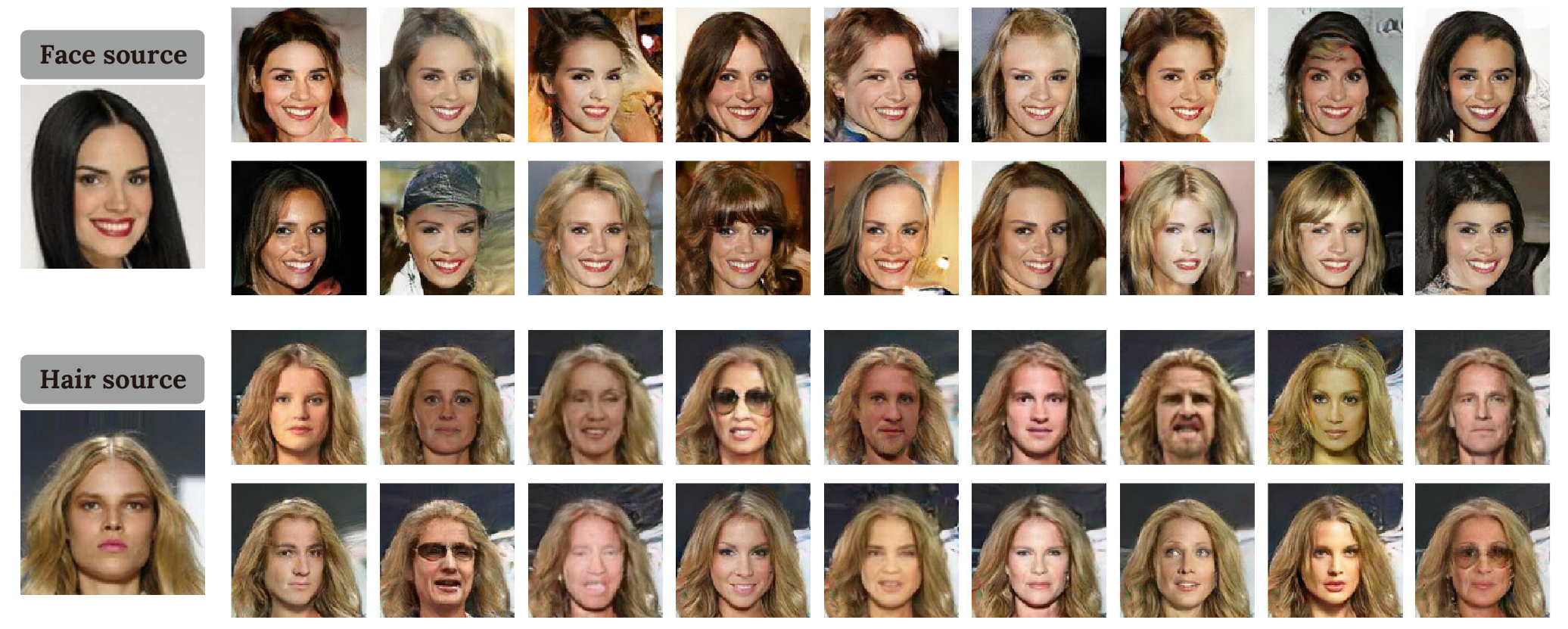} \\
    \vspace{5mm}
    \includegraphics[width=\linewidth]{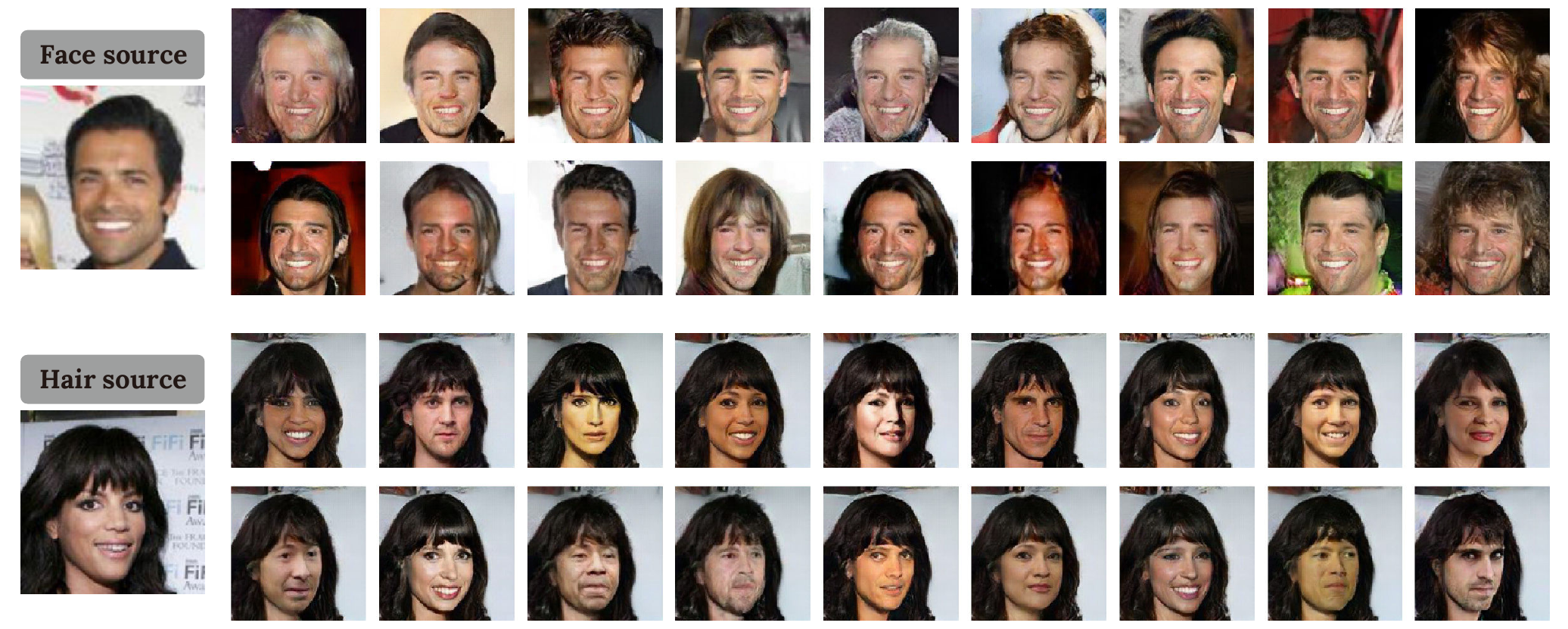}
    \caption{Additional results for random face parts sampling.}
    \label{fig:additional-random-sample}
\end{figure}

\begin{figure}[p]
    \centering
    \includegraphics[width=\linewidth]{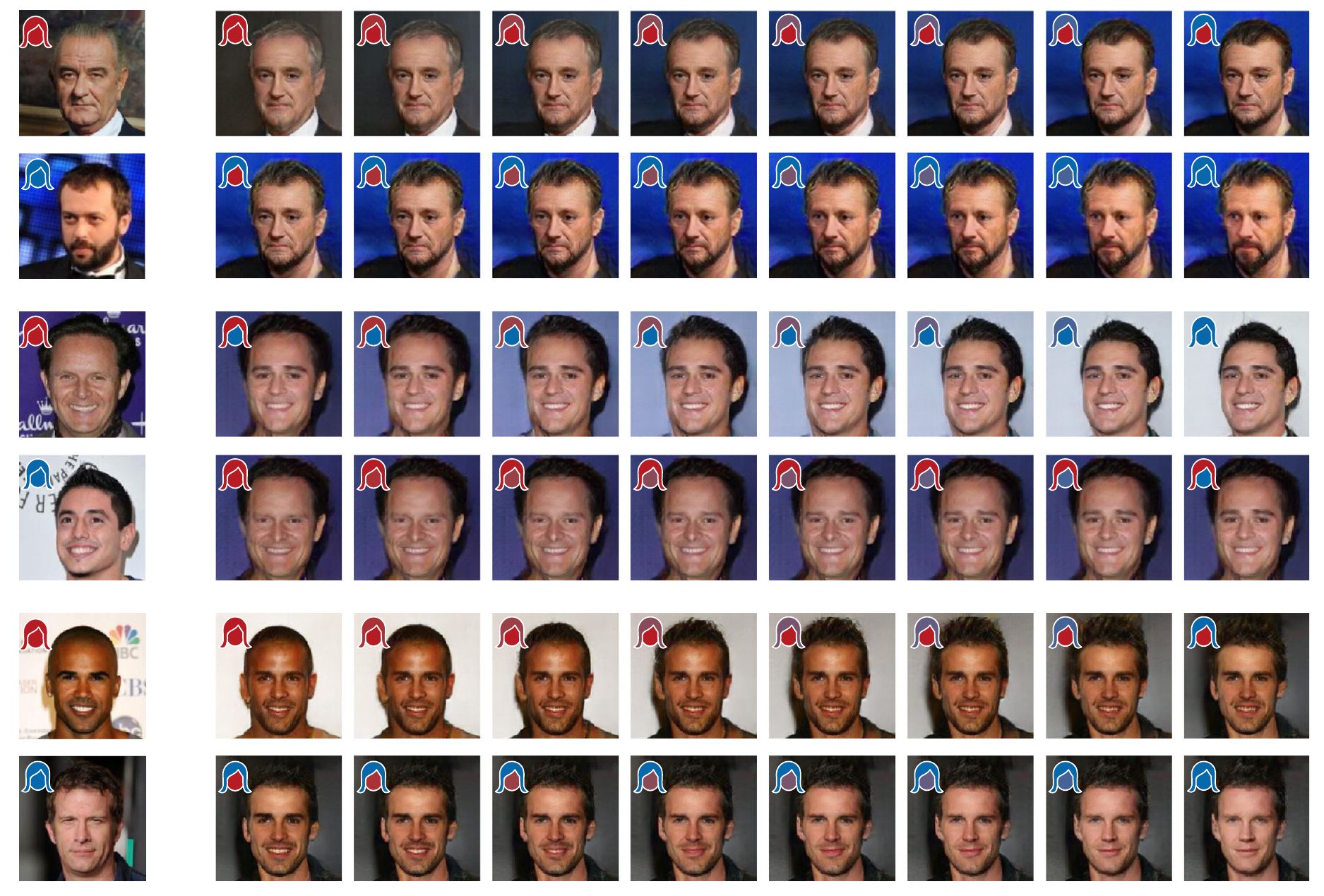} \\
    \vspace{5mm}
    \includegraphics[width=\linewidth]{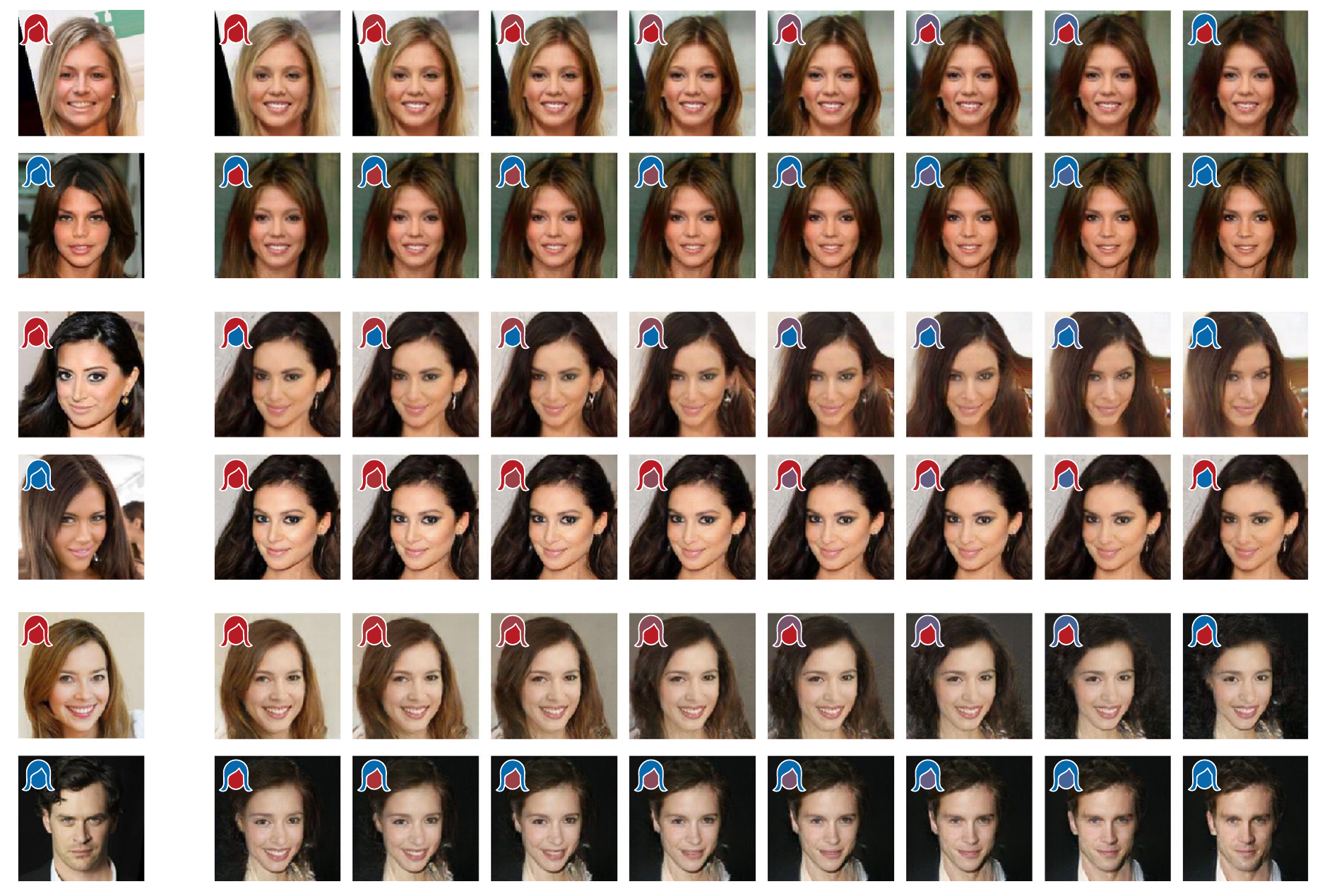}
    \caption{Additional results for face parts interpolation.}
    \label{fig:additional-interpolation}
\end{figure}

\begin{figure}[p]
    \centering
    \includegraphics[width=\linewidth]{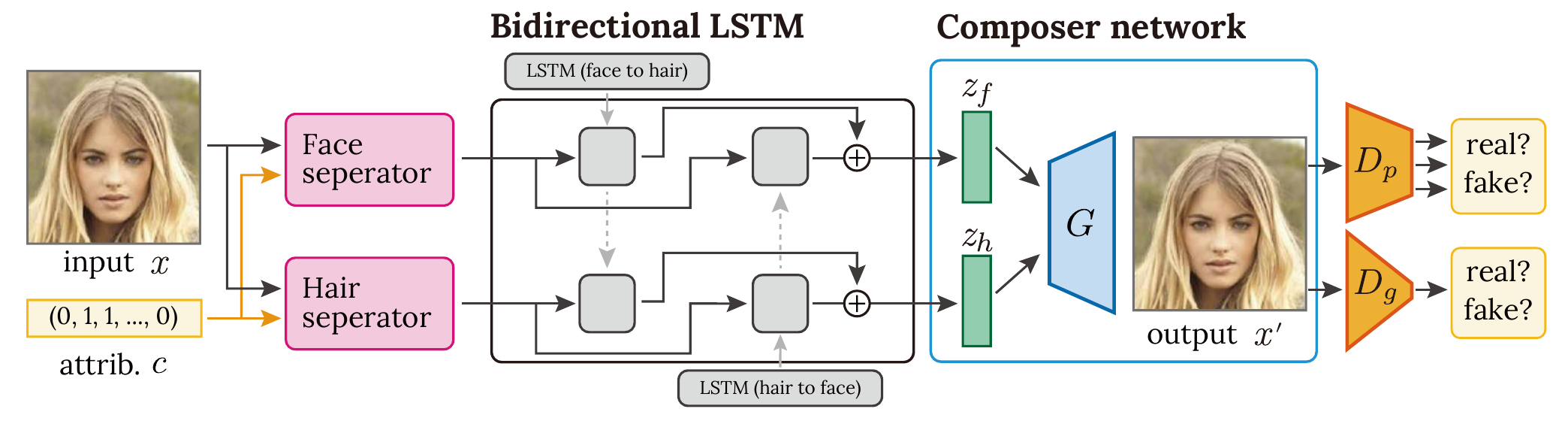}
    \caption{The network architecture of RSGAN with LSTM. In this architecture bidirectional LSTM is inserted after the two separator.}
    \label{fig:network-lstm}
\end{figure}

\begin{figure}[p]
    \centering
    \begin{tabular}{c}
        \textbf{RSGAN without LSTM} \\
        \includegraphics[width=\linewidth]{ResultsSwap} \\
        \\
        \\
        \textbf{RSGAN with LSTM} \\
        \includegraphics[width=\linewidth]{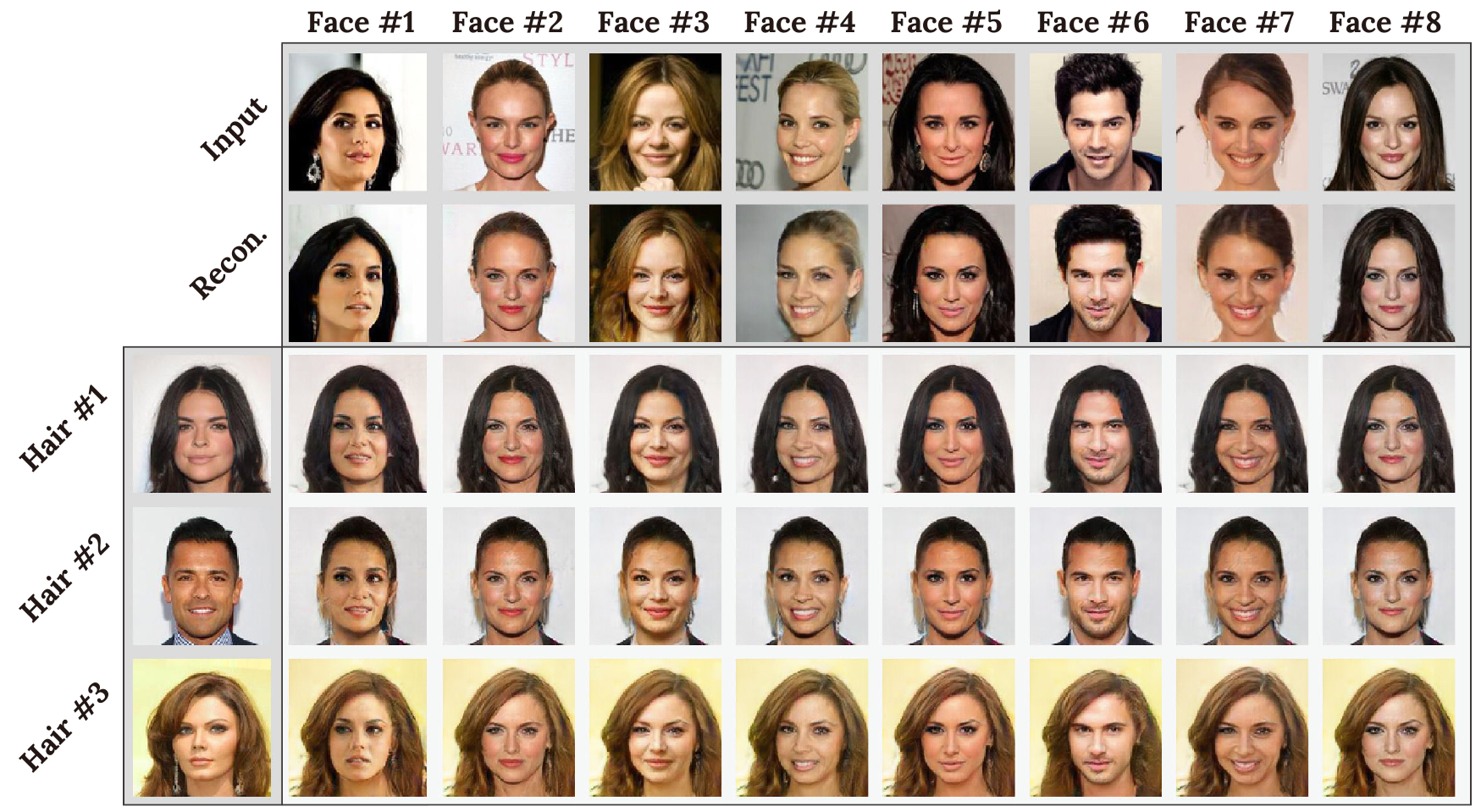} \\
    \end{tabular}
    \caption{Comparison of results with and without LSTM. The above image group illustrates the results of RSGAN without LSTM that is the same one as in \figref{fig:results-swapping}. The bottom group illustrates the results with LSTM.}
    \label{fig:compare-lstm}
\end{figure}

\begin{enumerate}
    \setlength\itemsep{1em}
    \item \textbf{RSGAN's training algorithm} (Algorithm~\ref{alg:training})
    \item \textbf{Additional results:}
    \begin{itemize}
        \setlength\itemsep{0.5em}
        \item Face swapping (\figref{fig:additional-face-swap})
        \item Visual attribute editing (\figref{fig:additional-attr-edit})
        \item Random face parts sampling (\figref{fig:additional-random-sample})
        \item Face parts interpolation (\figref{fig:additional-interpolation})
    \end{itemize}
    \item \textbf{RSGAN with LSTM}
    \begin{itemize}
        \setlength\itemsep{0.5em}
        \item Illustration of the network architecture with LSTM (\figref{fig:network-lstm})
        \item Face swapping results with LSTM (\figref{fig:compare-lstm})
    \end{itemize}
\end{enumerate}

\begin{algorithm}[p]
    \caption{Training pipeline of the proposed RSGAN.}
    \label{alg:training}
    \begin{algorithmic}[1]
        \Require $\lambda_{rec} = 4000$, $\lambda_{KL} = 1$, $\lambda_{adv \mathbar g} = 20$, $\lambda_{adv \mathbar p} = 30$, $\lambda_C = 1.0$, and $\lambda_{GC} = 50$.
        \While{Convergence not reached}
        \parState{\textbf{Step 1.} Compute loss functions:}
        \Indent
        \parState{Sample a batch $( x, x_f, x_h, c ) \sim P_{data}$ from the real data.}
        \parState{$z_{x_{f}} \leftarrow F_{E \mathbar x_{f}} (x, c),~z_{x_{h}} \leftarrow F_{E \mathbar x_{h}} (x, c)$.}
        \parState{Compute the KL losses, $\mathcal{L}_{KL \mathbar x_{f}}$ and $\mathcal{L}_{KL \mathbar x_{h}}$ with Eq.\,4.}
        \parState{$z_{c_{f}} \leftarrow F_{E \mathbar c_{f}} (c),~z_{c_{h}} \leftarrow F_{E \mathbar c_{h}}(c)$.}
        \parState{Compute the KL losses, $\mathcal{L}_{KL \mathbar c_{f}}$ and $\mathcal{L}_{KL \mathbar c_{h}}$ with Eq.\,4.}
        \parState{$x_f' \leftarrow F_{D \mathbar f} (z_{x_{f}}, z_{c_{f}}),~x_h' \leftarrow F_{D \mathbar h} (z_{x_{h}}, z_{c_{h}})$.}
        \parState{Compute the parts reconstruction losses, $\mathcal{L}_{rec \mathbar f}$ and $\mathcal{L}_{rec \mathbar h}$ with Eq.\,1 and Eq.\,2, respectively.}
        \parState{$x' \leftarrow G(z_{x_{f}}, z_{c_{f}}, z_{x_{h}}, z_{c_{h}})$.}
        \parState{Compute the reconstruction loss, $\mathcal{L}_{rec}$ with Eq.\,3.}
        \parState{Sample a batch of random vector $\hat{z}_{x_{f}}$, $\hat{z}_{x_{h}}$, $\hat{z}_{c_{f}}$, and $\hat{z}_{c_{h}}$ from the multi-variate standard normal distribution $P_{z}$.}
        \parState{$\hat{x}' \leftarrow G(\hat{z}_{x_{x_{f}}}, \hat{z}_{c_{f}}, \hat{z}_{x_{h}}, \hat{z}_{c_{h}})$.}
        \parState{Compute the adversarial losses, $\mathcal{L}_{adv \mathbar g}$ and  $\mathcal{L}_{adv \mathbar p}$ with Eq.\,5.}
        \parState{Estimate visual attributes $c^{*}$, $c'$ and $\hat{c}'$ using the classifier network.}
        \parState{Compute the classification loss, $\mathcal{L}_{C}$ with Eq.\,6.}
        \parState{Compute the classification losses for the composer network, $\mathcal{L}_{GC}$ with Eq.\,7.}
        \EndIndent
        
        \parState{\textbf{Step 2.} Update network parameters $\Theta$:}
        \Indent
        \parState{$\mathcal{L}_{G} \inclongleftarrow \mathcal{L}_{adv \mathbar g} + \mathcal{L}_{adv \mathbar p} + \lambda_{GC} \mathcal{L}_{GC}$}
        \parState{$\Theta_{F_{E \mathbar x_{f}}} \inclongleftarrow -\nabla_{\Theta_{F_{E \mathbar x_{f}}}} (\lambda_{rec} \mathcal{L}_{rec \mathbar f} + \lambda_{KL} \mathcal{L}_{KL \mathbar x_{f}} + \mathcal{L}_{G})$}
        \parState{$\Theta_{F_{E \mathbar x_{h}}} \inclongleftarrow -\nabla_{\Theta_{F_{E \mathbar x_{h}}}} (\lambda_{rec} \mathcal{L}_{rec \mathbar h} + \lambda_{KL} \mathcal{L}_{KL \mathbar x_{h}} + \mathcal{L}_{G})$}
        \parState{$\Theta_{F_{E \mathbar c_{f}}} \inclongleftarrow -\nabla_{\theta_{F_{E \mathbar c_{f}}}} (\lambda_{rec} \mathcal{L}_{rec \mathbar f} + \lambda_{KL} \mathcal{L}_{KL \mathbar c_{f}} + \mathcal{L}_{G})$}
        \parState{$\Theta_{F_{E \mathbar c_{h}}} \inclongleftarrow -\nabla_{\Theta_{F_{E \mathbar c_{h}}}} (\lambda_{rec} \mathcal{L}_{rec \mathbar h} + \lambda_{KL} \mathcal{L}_{KL \mathbar c_{h}} + \mathcal{L}_{G})$}
        \parState{$\Theta_{F_{D \mathbar f}} \inclongleftarrow -\nabla_{\Theta_{F_{D \mathbar f}}} (\lambda_{rec} \mathcal{L}_{rec \mathbar f})$}
        \parState{$\Theta_{F_{D \mathbar h}} \inclongleftarrow -\nabla_{\Theta_{F_{D \mathbar h}}} (\lambda_{rec} \mathcal{L}_{rec \mathbar h})$}
        \parState{$\Theta_{G} \inclongleftarrow -\nabla_{\Theta_{G}} (\lambda_{rec} \mathcal{L}_{rec} + \mathcal{L}_{adv})$}
        \parState{$\Theta_{D_g} \inclongleftarrow -\nabla_{\Theta_{D_g}} (\lambda_{adv \mathbar g} \mathcal{L}_{adv \mathbar g})$}
        \parState{$\Theta_{D_p} \inclongleftarrow -\nabla_{\Theta_{D_p}} (\lambda_{adv \mathbar p} \mathcal{L}_{adv \mathbar p})$}
        \parState{$\Theta_{C} \inclongleftarrow -\nabla_{\Theta_{C}} (\lambda_{C} \mathcal{L}_{C})$}
        \EndIndent
        \EndWhile
    \end{algorithmic}
\end{algorithm}

\end{document}